\documentclass[11pt]{article}

\usepackage{graphicx} 
\usepackage[utf8]{inputenc}
\usepackage{booktabs}
\usepackage{lscape}
\usepackage[top=2.5cm, bottom=2.5cm, left=3.5cm, right=3.5cm]{geometry}
\usepackage[T1]{fontenc}
\usepackage{setspace}
\usepackage{caption}
\usepackage{chngcntr}
\usepackage[round]{natbib}
\usepackage[english]{babel}
\usepackage{amsmath, amsfonts, amssymb, amsthm}
\usepackage{supertabular}
\usepackage{longtable}
\usepackage{scrextend}
\usepackage{xcolor}
\usepackage{multirow}
\usepackage{algorithm}
\usepackage{algpseudocode}
\usepackage{array, tabularx, caption, boldline}
\usepackage{graphicx}
\usepackage{cellspace}
\usepackage{lscape}
\usepackage{array}
\newcolumntype{P}[1]{>{\centering\arraybackslash}p{#1}}
\usepackage{array, makecell}
\usepackage{lineno}
\usepackage[hidelinks]{hyperref}
\usepackage{colortbl}
\modulolinenumbers[5]

\newtheorem{definition}{Definition}%

\newcommand\independent{\protect\mathpalette{\protect\independenT}{\perp}}
\def\independenT#1#2{\mathrel{\rlap{$#1#2$}\mkern2mu{#1#2}}}

\usepackage{thmbox}

\usepackage{refcount}
\usepackage{tikzpagenodes}

\newcommand\dcorr{%
}

\newcommand\fcorr{%
}

\begin{document}

\title{Measuring and Mitigating Biases\\in Motor Insurance Pricing}

\author{Mulah Moriah$^{1,2}$, Franck Vermet$^{1,2}$ \& Arthur Charpentier$^3$}
\date{%
    $^1$Euro-Institut d'Actuariat Jean Dieudonné (EURIA)\\ $^2$Univ Brest, CNRS,  UMR 6205, Laboratoire de Math\'ematiques de Bretagne Atlantique, France \\
    $^3$Universit\'e du Qu\'ebec \`a Montr\'eal
}

\maketitle

\begin{abstract}
\singlespace The non-life insurance sector operates within a highly competitive and tightly regulated framework, confronting a pivotal juncture in the formulation of pricing strategies. Insurers are compelled to harness a range of statistical methodologies and available data to construct optimal pricing structures that align with the overarching corporate strategy while accommodating the dynamics of market competition.
\dcorr Given the fundamental societal role played by insurance, premium rates are subject to rigorous scrutiny by regulatory authorities. Consequently, the act of pricing transcends mere statistical calculations and carries the weight of strategic and societal factors. These multifaceted concerns may drive insurers to establish equitable premiums, considering various variables. For instance, regulations mandate the provision of equitable premiums, considering factors such as policyholder gender. Or mutualist groups in accordance with respective corporate strategies can implement age-based premium fairness. In certain insurance domains, the presence of serious illnesses or disabilities are emerging as new dimensions for evaluating fairness.
Regardless of the motivating factor prompting an insurer to adopt fairer pricing strategies for a specific variable, the insurer must possess the capability to define, measure, and ultimately mitigate any fairness biases inherent in its pricing practices while upholding standards of consistency and performance. This study seeks to provide a comprehensive set of tools for these endeavors and assess their effectiveness through practical application in the context of automobile insurance. Results show that fairness bias can be found in historical data and models, and that fairer outcomes can be obtained by more fairness-aware approaches.\fcorr

\end{abstract}


\section{Introduction}\label{sec1}

\subsection{Motivation}

The insurance industry is characterized by an inherent reversal in its production cycle, where insurers request a fixed premium at the time of policy subscription in exchange for coverage against uncertain risks in terms of both occurrence and magnitude. This inversion underscores the statistical nature that envelops the pricing of insurance, necessitating adherence to statistical theory for the estimation and coverage of random events. Beyond these statistical considerations, insurance premiums represent the equitable price for insurance services, encompassing a multitude of strategic challenges. The insurance market is increasingly competitive, comprising established incumbents and more agile newcomers striving to gain market share. This competitive landscape is further catalyzed by continuously evolving regulations aimed at fostering competition among industry players.

Hence, participants in the insurance market must offer competitive pricing strategies that align with their corporate strategies and communication, while also employing tools tailored to diverse distribution processes.

Recent years have witnessed the widespread adoption of machine learning algorithms, neural networks, and the utilization of vast datasets. This adoption is attributable to scientific advancements, increased computational power, enhanced accessibility to technology, and the proliferation of data. These emerging technologies have made significant inroads into various business sectors, including the insurance industry.

These data and algorithms serve as decision support tools, aiding in policyholder segmentation, risk comprehension, and the consideration of various factors associated with it. Therefore, industry stakeholders must incorporate these new elements to maintain their competitiveness. Nevertheless, the extensive use of massive data and intricate algorithms has brought the issue of transparency to the forefront. It is imperative that premiums and decisions are both explainable and fair, given the high stakes involved. This is not simply about algorithms assisting in trivial choices like movie selection; rather, it involves algorithms determining the cost of access to insurance services for various population segments. Concerns regarding fairness and ethics have been integral to our societies and philosophies for centuries. Although subject to interpretation, fairness can be defined as the ability to place individuals on an equal footing while acknowledging the differences that exist among them.

In response to these considerations, regulatory measures such as the gender directive have been implemented to promote fairness. Reforms related to access to borrower insurance also represent a form of fairness enforcement. Consequently, insurance industry participants may need to construct fairer premiums with regard to so-called sensitive or protected variables. These fairness constraints can emanate from regulatory requirements, as seen in the case of gender, or be driven by commercial and strategic objectives, such as those related to age in specific companies.

In the context of this study, bias is understood as a form of discrimination, \textbf{signifying the undesirable impact of a sensitive variable on a variable of interest}. \dcorr For instance, this could include the effect of gender on insurance premiums within the framework established by the gender directive (\cite{genderdir_2008}). Since 2016, numerous research studies (\cite{angwin2016chaos,chouldechova2017disparate})\fcorr have identified instances of discrimination in decision-making tools across various domains, such as the risk assessment tool for recidivism in the United States, Google Images algorithms, Amazon's application processing algorithms, and lending and financing algorithms, among others. Historically, the solution has often been to circumvent the issue rather than address it from an ethical perspective.

One simplified scenario that can be examined involves the use of a gender-correlated variable to determine premiums. In this case, gender is not explicitly factored into the estimates, resulting in fairness by omission. However, the presence of a correlation between gender and the non-sensitive explanatory variable leads to an indirect effect of gender on the estimated premiums. Even though gender is seemingly reprocessed or removed from the model, its association with other variables allows its influence on the target variable to persist. This is because adjustments only account for the direct impact of sensitive variables and not their indirect effects, despite these variables exerting a significant influence on the distributions of other explanatory variables. Factors such as age, gender, and disability influence choices related to activities, risk tolerance, product preferences, and so on. Consequently, it is imperative to first establish the means to define and measure the fairness of the constructed models and subsequently mitigate this bias through fairness-aware approaches.

\dcorr
\subsection{Agenda}

The purpose of this article is to provide actuaries with tools to understand, measure and mitigate the unwanted effect of a sensitive variable in a pricing problem.\fcorr In Section \ref{sec:2}, we will discuss fairness notions and statistical measures of fairness. The bigger picture will be shown while focusing on important elements for pricing. Then, in Section \ref{sec:bias_mitigation}, we will present methods that can be used to reduce fairness bias in pricing models. Finally, in Sections \ref{sec:3} and \ref{sec:5}, we present the results of fairness implementation on a real car insurance pricing, respectively with measures of biases (in Section \ref{sec:3}) and a description of the impact of mitigation  (in Section \ref{sec:5}). Note that all implementations are made in Python 3.8.

\subsection{Notations}

Throughout this document, the variable denoted as $Y$ is the target variable we aim to predict. This variable can take on either categorical or quantitative values (discrete or continuous, but positive). Given the primary focus on pricing, $Y$ predominantly assumes a quantitative nature. However, for the sake of illustration, it might be temporarily treated as a binary variable. In the context of addressing this supervised machine learning problem, we introduce an algorithm denoted as $m$, a set of non-sensitive features represented by the vector $\boldsymbol{X} = (X_{1}, X_{2}, \dots, X_{p})$, and sensitive attributes denoted as $\boldsymbol{S}$. The vector $\boldsymbol{S}$ comprises variables that are intended to have no influence on the models, either intentionally or inadvertently. These variables are typically discrete, particularly in the context of examining fairness in pricing, and may explicitly pertain to attributes such as gender. If a single sensitive attribute is considered, as it will mostly be the case in this document, notation $S$ will be used (see \cite{charpentier2023sequential} for a detailed discussion about multiple sensitive attributes). Furthermore, we assume that $S$ is a binary sensitive variable, with possible values of $0$ and $1$, which may correspond to the gender of the policyholder.

Each data point in our dataset, which can represent individuals, contracts, or claims, is identified as $(y_i, \boldsymbol{x}_i, s_i)$, with $i$ ranging from $1$ to $n$, and $n_0$ and $n_1$ signifying the numbers of observations where $s_i=0$ and $s_i=1,$ respectively. Note that lowercase letter variables denote observations, whereas uppercase letter variables denote random variables.

We introduce $\widehat{y} = \widehat{m}(\boldsymbol{x})$ as the predictions generated by an "unfair model" (or "unaware model", as defined by \cite{dwork2011aware, dwork2012fairness}), and $\widetilde{y} = \widetilde{m}(\boldsymbol{x})$ as the predictions generated by a fair model. \dcorr $\widetilde{m}$ is a model that takes into account defined fairness constraints to minimize the effects of $s$ on $y$. As studied in section \ref{sec:bias_mitigation}, these constraints can take the form of criteria for variable selection (preprocessing constraints), penalized learning (in-processing approach) and output correction (post-processing approach).  We can then define individual fairness bias, denoted as $\varepsilon$, as the difference between these two predictions, i.e., $\varepsilon = \widehat{y} - \widetilde{y}$.\fcorr

We also introduce the function $\mathcal{V}^s(\boldsymbol{x}_i)$, which represents the $k$-nearest neighbors in the group $S=s$ associated with the $i$-th individual.

\section{Measuring fairness and biases}\label{sec:2}

While economists have engaged in discussions on discrimination for several decades, as evidenced in works such as \cite{edgeworth1922equal}, \cite{becker2010economics}, and \cite{phelps1972statistical}, recent contributions from the field of computer science have sought to formalize key fairness concepts and the notion of non-discrimination with respect to specific sensitive attributes denoted as "S". These discussions primarily pertain to various classifiers denoted as "m", and are exemplified in works such as \cite{dwork2012fairness}, \cite{hardt2016equality}, \cite{berk2017convex}, and \cite{corbettdavies2017algorithmic}. For an overview of state-of-the-art developments in insurance models, please refer to \cite{charpentier2023springer}.

As the literature in this area is relatively recent, references and consensus have not yet fully coalesced, as noted by \cite{angwin2016chaos}: "{\em The rapid growth of this emerging field has led to highly inconsistent motivations, terminologies, and notations, posing a significant challenge in cataloging and comparing definitions}." Furthermore, \cite{castel2022clarif} describe the multiplicity of fairness definitions as a "zoo of definitions," remarking, "{\em The researcher or practitioner approaching this facet of machine learning for the first time can easily feel confused and somewhat lost in this maze of definitions. These various definitions capture different facets of the fairness concept, but to the best of our knowledge, a comprehensive understanding of the broader landscape where these measures reside remains elusive.}"

\dcorr
Fairness is intuitively understood as the absence of any association between the sensitive variable and the variable of interest. Fair predictions $ \widetilde{y}$ will then be independent of $S$. Indeed, independence between these variables implies the absence of any direct or indirect relationships, thereby precluding the existence of fairness bias. However, this is just one facet of the fairness concept.\fcorr

In this section, we will introduce the two primary types of fairness, namely group fairness and individual fairness, with group fairness being the more prevalent of the two. 

\subsection{Group or Statistical Fairness}
\dcorr
Group fairness necessitates equality of treatment among groups based on the sensitive variables. For that, statistical properties are specified using the model $m$ and $\widehat{Y}$ to compare each group. The overarching objective is to ensure that individuals from both privileged and unprivileged groups are treated equitably in terms of the specified statistical properties.

Fairness was initially introduced using conditional probabilities in the context of binary classification (see \cite{usem1978}).\fcorr This concept can be extended to scenarios where the variable $Y$ is non-binary, incorporating moment properties, a weak version, or distribution properties, a strong version. These can be linked to correlation properties and independence conditions, respectively.

\paragraph{Independence (demographic parity):} States that predictions ($\widehat Y$) should not rely on the sensitive variable $S$. It emphasizes that predictions must be independent of $S$: 
$$\widehat Y \independent S.$$
This principle highlights the need for predictions to be unrelated to $S$, without mentioning the target variable ($Y$). In practical scenarios like insurance pricing, if $S$ is age and $Y$ is premium, it suggests premiums should be consistent across age groups. However, enforcing this fairness principle may seem counterintuitive if certain age groups pose higher risks, implying less restrictive models to ensure consistent premiums regardless of age. This approach may contradict traditional fairness perceptions, which prioritize equal treatment regardless of sensitive attributes. This fairness principle becomes crucial when there's concern about unfair information in the target variable ($Y$), especially due to historical bias in the data, reflecting past unfair behaviors or decisions. In such cases, emphasizing fairness regarding $S$ may be necessary for corrective measures.

\dcorr
\paragraph{Separation (equalized odds):} states that predictions ($\widehat Y$) should be independent of the sensitive variable ($S$) when the true value of the target variable ($Y$) is known. This means that once the actual outcome ($Y$) is revealed, the predictions should not be influenced by the sensitive attribute ($S$):
$$\widehat Y \independent S ~\vert~ Y.$$
 In the context of separation, any disparities in treatment between groups based on the sensitive attribute ($S$) must be justifiable by the actual value of the target variable ($Y$). For example, in a scenario involving premium and age, premiums may vary for each age group, based on risk factors independent of age. This approach reduces the influence of the sensitive attribute ($S$) while preserving valuable information in the target variable ($Y$). However, it's only feasible when $Y$ isn't affected by historical bias. If unfair information is present in $Y$, it will easily affect the predictions $\widehat Y$.
 
\fcorr

\paragraph{Sufficiency (predictive parity):}
examines fairness regarding the target variable ($Y$). It aims for independence between $Y$ and the sensitive attribute ($S$), given the predictions ($\widehat Y$): 
$$Y \independent S ~\vert~ \widehat Y.$$
This approach doesn't require knowledge of the true $Y$ for unseen individuals, addressing fairness directly. In feature selection and modeling, $Y$ often faces selection issues due to observed data limitations. By focusing on $\widehat Y$ as a starting point, this approach helps mitigate such issues associated with $Y$.

Different scopes of group fairness are inherently incompatible. Studies show simultaneous fulfillment of multiple fairness criteria is challenging, except in special cases. References like \cite{chouldechova2017disparate}, \cite{kleinberg2016inherent}, \cite{berk2018criminal}, and \cite{charpentier2023springer} explore this. This understanding leads to the introduction of the following definition of statistical fairness.

\begin{definition}[Statistical (Group) Fairness]
Let $\widehat{y}={m}(\boldsymbol{x})$.
A model $m$ is classified as strongly fair if it satisfies the following conditions:
$$
\begin{cases}
    \text{demographic parity}:&~(\widehat{Y}\vert S=s) \overset{\mathcal{L}}{=} \widehat{Y},~\forall s\\
    \text{equalized odds}:&~(\widehat{Y}\vert S=s,Y) \overset{\mathcal{L}}{=} (\widehat{Y}\vert Y),~\forall s\\
    \text{predictive parity}:&~(Y\vert S=s,\widehat{Y}) \overset{\mathcal{L}}{=} (Y\vert \widehat{Y}),~\forall s
\end{cases}$$
while $m$ is considered weakly fair if it fulfills the following conditions:
$$
\begin{cases}
    \text{demographic parity}:&~\mathbb{E}[\widehat{Y}\vert S=s] = \mathbb{E}[\widehat{Y}],~\forall s\\
    \text{equalized odds}:&~\mathbb{E}[\widehat{Y}\vert S=s,Y] = \mathbb{E}[\widehat{Y}\vert Y],~\forall s\\
    \text{predictive parity}:&~\mathbb{E}[Y\vert S=s,\widehat{Y}] = \mathbb{E}[Y\vert \widehat{Y}],~\forall s
\end{cases}$$
\end{definition}
These definitions provide a framework for categorizing models as either strongly fair or weakly fair based on their compliance with different fairness criteria, namely demographic parity, equalized odds, and predictive parity.

\subsection{Individual fairness}
Individual fairness, also known as similarity-based fairness, asserts that similar individuals should receive similar predictions. Unlike group fairness, which aggregates outcomes at the group level, individual fairness compares individuals directly. This concept is formalized as "disparate treatment":

\begin{definition}[Disparate Treatment]
A model $m$ exhibits disparate treatment if, given the explanatory variables $\boldsymbol{X}$, the predictions $\widehat Y$ and the sensitive attribute $S$ are dependent. To achieve fairness:
$$\widehat Y  \independent S \vert \boldsymbol{X}.$$
\end{definition}

Individual fairness and disparate treatment embody an intuitive conception of fairness, seen in scenarios like car insurance pricing, where identical individuals with different sensitive attribute values receive the same predictions. However, this approach overlooks relationships between explanatory variables ($\boldsymbol{X}$) and $S$, impacting fairness enforcement.

The challenge of individual fairness arises from sensitive variables significantly affecting other variables' distribution. Attributes like gender, age, disabilities, ethnicity, and illnesses shape habits, preferences, and behaviors, complicating defining "similar individuals."

To address these questions, \cite{dwork2011aware} introduced the Lipschitz constraint concept:
$$d_{Y}(\widehat{y}_{i}, \widehat{y}_{j}) < \lambda d_{\boldsymbol{X}}(\boldsymbol{x}_{i},\boldsymbol{x}_{j}),$$
Here, $d_{Y}$ and $d_{\boldsymbol{X}}$ measure distances in target and explanatory variable spaces, respectively. Similarity in the $Y$ space, like between premiums, can be defined using metrics such as $\vert\widehat y_i - \widehat y_j\vert$ or $(\widehat y_i - \widehat y_j)^2$. 
\dcorr
The focus is on defining individuals as similar when they have different values of the sensitive attribute ($S$). And, disparate treatment will mean that regardless of the value of $S$, if two individuals are similar on $\boldsymbol{X}$, they must be similar on $Y$ and thus have the same premium.
\fcorr

Despite its intuitive appeal, individual fairness is less frequently used than group fairness, largely due to addressing various interactions between variables' complexity. Causality-based criteria may offer a promising avenue, but practical implementation remains challenging. These causal models are not discussed here (for more information, refer to \cite{kusner2017counterfactual}, \cite{Carey2022}, and \cite{Pearl1998}).

\subsection{Quantifying unfairness}

\dcorr In light of the definitions provided earlier, it is imperative that fairness metrics possess the ability to quantify full/conditioned independence in the context of group fairness and also discern discrepancies in individual predictions, contingent upon the proximity of individuals in the case of individual fairness.\fcorr

\subsubsection{From the binary to the general case}\label{label:kendall:HGR}

In the case where both the target variable ($y$) and the sensitive attribute ($s$) are binary variables, metrics can be defined using the confusion matrix of $\widehat{y}$ and $s$. For instance, the disparate impact metric can be derived as:
$$\frac{\mathbb{P}( \widehat{Y}=1 \vert S=1)}{\mathbb{P}( \widehat{Y}=1 \vert S=0)}.$$
In the context of demographic parity, when this metric is closer to one, it indicates a fairer model. Similarly, another fairness metric, denoted as $M$, is defined as:
\begin{eqnarray*}
M_1 &=& |{\mathbb{P}( \widehat{Y}=1 \vert S=1)}-{\mathbb{P}( \widehat{Y}=1 \vert S=0)}|  \\
 &=& |{\mathbb{P}( \widehat{Y}=0 \vert S=1)}-{\mathbb{P}( \widehat{Y}=0 \vert S=0)}| ~=~ M_0.
\end{eqnarray*}
In the case of demographic parity, the closer the value of $M_1$ (or $M_0$) is to zero, the fairer the model.

To address separation and sufficiency, a confusion matrix involving $\widehat{Y}$, $Y$, and $S$ can be constructed, enabling the definition of metrics such as true positive and false negative rates.
To measure fairness in accordance with the equalized odds definition, the disparate mistreatment can be calculated as follows:
$$\begin{cases}
M_{1|0} = \vert\mathbb{P}(\widehat{Y} = 1\vert Y=0, S = 1) - \mathbb{P}(\widehat{Y} = 1 \vert Y=0,S =0)\vert, \\
M_{0|1} = \vert \mathbb{P}(\widehat{Y} = 0 \vert Y=1, S = 1) - \mathbb{P}(\widehat{Y} = 0 \vert Y=1,S =0)\vert.    
\end{cases}
$$
In the context of equalized odds, the closer these values are to zero, the fairer the model is considered to be.

In a more general context involving non-binary variables, various correlation-based metrics, such as Kendall's tau, Pearson's correlation, and Spearman's rho, can be employed to assess dependencies between variables. \dcorr However, in the pursuit of capturing all forms of dependence in the relationship between $\widehat Y$ and $S$, the Hirschfeld-Gebelein-Renyi ($\mathrm{HGR}$) maximal correlation measure appears to outperform other methods.\fcorr 

The $\mathrm{HGR}$ maximal correlation, originally introduced by \cite{hirschfeld1935} and further developed by \cite{Gebelein1941} and \cite{renyi1959dep}, offers the ability to quantify both linear and nonlinear relationships while adhering to essential properties for a reliable measure of dependency, as established by \cite{renyi1959dep}. It enjoys broad acceptance within the field of statistics.
For two continuous or discrete variables, the $\mathrm{HGR}$ maximal correlation is precisely equal to zero if and only if the variables are independent. This makes it a valuable tool for detecting and quantifying dependencies between variables.

\begin{definition}[$\mathrm{HGR}$]
For two random variables $U$ and $V$ respectively with values in $\mathcal{U}$ and $\mathcal{V}$, 

$$\mathrm{HGR}(U,V) = \max_{f\in\mathcal{F}_\mathcal{U} ,\ g\in\mathcal{G}_\mathcal{V} }\mathbb{E}[f(U)g(V)],$$
where
$$
\begin{cases}
\mathcal{F}_\mathcal{U}  = \lbrace f:\mathcal{U} \to \mathbb{R}:\mathbb{E}[f(U)] = 0\text{ and }\mathbb{E}[f^2(U)] = 1\rbrace \\
\mathcal{G}_\mathcal{V}  = \lbrace g:\mathcal{V} \to \mathbb{R}:\mathbb{E}[g(V)] = 0\text{ and }\mathbb{E}[g^2(V)] = 1\rbrace \\
\end{cases}
$$
In our case, $U$ will be the sensitive attribute, and $V$ the premium (or one of its components). 
\end{definition}

In the realm of fairness quantification, \cite{mary2019fairness} proposed the use of the $\mathrm{HGR}$ maximal correlation measure, and \cite{grary2022} explored its application in the insurance domain. However, obtaining the exact value of this metric is not straightforward, as it is defined in infinite-dimensional function spaces. \dcorr Several authors have used Witsenhausen’s linear algebra characterization (see \cite{Witsenhausen1975cost}) to compute the HGR. This measure can also be related to Kernel Based Nonlinear Canonical Analysis \cite{darolles2004kernel}. Therefore, \cite{mary2019fairness} combine kernel density estimation (KDE) (specifically a gaussian kernel) to these characterizations to obtain an approximation of the HGR, called HGR\_KDE. \fcorr Importantly, this estimator has been demonstrated to preserve the fundamental properties of the original $\mathrm{HGR}$ measure and delivers strong performance. Therefore, in this paper, one of the  metrics utilized is the $\mathrm{HGR}$ maximal correlation, and we employ the implementation through HGR\_KDE, as recommended by its authors.

\subsubsection{Focus on the case of pricing}\label{sec:2:3:2}

In pricing, where the target variable $Y$ is continuous, confusion matrix metrics are irrelevant. Various correlation and distribution-based metrics related to the sensitive attribute $S$ become relevant. Here, we define $S$ as a binary variable representing gender.

One commonly used metric is Kendall's tau. Additionally, $\mathrm{HGR}$ estimation is valuable and applicable to all variables.

\dcorr
Probabilistic distances and divergences can be computed on conditional distributions $Y \vert S$. For instance, Kullback–Leibler's divergence (KL) assesses disparities between premiums for men and women.
\fcorr

The $p$-value from the Kolmogorov-Smirnov (KS) test is employed for fairness evaluation. This test complements divergences, being particularly sensitive to extreme values. These metrics are implemented using the Python package \texttt{Scipy}. The KL divergence is calculated using relative entropy, following the approach proposed in \cite{boyd2004convexopt}.

The concept of individual fairness with continuous variables has received less attention. Noteworthy contributions include \cite{dwork2011aware}, who proposed using $k$-nearest-neighbors for proximity assessment.

\paragraph{Adaptation flip-test} 
Inspired by \cite{black2020fliptest} and \cite{dwork2011aware}, we adapt a metric for continuous variables to define individual fairness. We assume a distance metric or algorithm effectively measures proximity between individuals.

\begin{itemize}
    \item Select individuals within group $s_{i}=1$.
    \item Find their $k$ nearest neighbors of the opposite gender.
    \item Calculate differences between their predictions $\widehat{y}_i$ and the average predictions of their neighbors $\mathcal{V}^0(\boldsymbol{x}_i)$ :
    $$\Delta_{\mathcal{V}}^{i} = \widehat{y}_{i} - \frac{1}{k} \sum_{x_{j}\in \mathcal{V}^0(\boldsymbol{x}_i)}\widehat{y}_{j}.$$
    \item Average these distances :
     $$\Tilde{FT}_{1} = \frac{1}{n_1}\sum^{n_1}_{i=1} \Delta_{\mathcal{V}}^{i}.$$
\end{itemize}

This process is repeated starting with $s_i=0$ to obtain $\Tilde{FT}_{0}$. A value closer to zero indicates fairer model.

However, it's crucial to note that the quality of the proximity metric provided by the $k$-nearest-neighbors model is the primary limitation. To address this, we conduct hyperparameter tuning and variable selection using a grid search approach on our model. This optimization involves selecting an appropriate distance metric and its parameter settings, along with determining the optimal number of neighbors. Additionally, we optimize the number and specific variables used in model construction. The objective is to identify the $k$-nearest-neighbors model that minimizes individual distance while minimizing differences in premiums. We perform a grid search on these parameters and validate the results using a test dataset. This approach offers a satisfactory solution with significantly lower operational costs compared to a causal study. Minimizing bias and distance helps ensure that the presented bias level isn't artificially inflated by the model, aligning with the concept of counterfactual fairness discussed in \cite{kusner2017counterfactual}, \cite{de2021transport}, and more recently \cite{charpentier2023transport}.

Once bias is detected, it's crucial to attempt mitigation while preserving the performance and consistency of the models constructed.

\section{Bias mitigation}\label{sec:bias_mitigation}

Fairness bias mitigation involves reducing the unwanted influence of the sensitive variable and the estimated variable of interest in models, data, or results. There is no consensus on bias mitigation approaches, as they are closely tied to specific use cases. Binary $y$ is often preferred in the literature, but in this study, we present methods applicable to regression and adapted for insurance contexts. Mitigation can occur before, during, or after modeling.

Incorporating fairness into the modeling process typically negatively impacts performance. Many papers have observed performance drops, with attempts to quantify these reductions \citep{delbarrio2020review}. Therefore, maintaining acceptable performance while enforcing fairness is crucial.

We evaluate model performance using two mean metrics: Root Mean Square Error (RMSE) and losses over premiums ratio (LR):
\begin{equation}\label{RMSE_LossRatio}
\mathrm{RMSE} = \sqrt{\frac{1}{n}\sum_{i=1}^{n} w_i(\widehat{y}_i - y_i)^2},
\ \mathrm{LR} = \frac{\sum_{i=1}^{n} {y}_i}{\sum_{i=1}^{n} \widehat{y}_i},
\end{equation}
where $w_i$ may represent a weight based on the variable $y$ considered. To balance performance and fairness metrics, we use them to guide optimization and comparison. A scenario is non-dominated when no other model achieves better performance and fairness simultaneously. Dominated scenarios can be surpassed by others in terms of fairness and performance.

Non-dominated scenarios warrant further investigation, and choosing the best dominant scenario depends on decision-maker constraints. Some prioritize fairness attainment, while others focus on maintaining performance levels.

We will present various methods in the following sections and summarize their advantages and disadvantages in Table 
 \ref{tab:pro_cons}.

\subsection{Pre-processing mitigation} 

These mitigation methods involve data transformations aimed at reducing bias while retaining relevant information. The methods we have implemented include: total removal of variables correlated with the sensitive variable (Section \ref{sec:3:1:1}), removal of linear correlations (Section \ref{sec:correlation}) and an adaptation of the fair-SMOTE method (Section \ref{sec:fair:smote}) proposed by \cite{Chakraborty2021smote}. Originally designed for binary cases, we have modified it to suit the needs of bias mitigation in non-life insurance pricing.



\subsubsection{Total deletion}\label{sec:3:1:1}
\dcorr This straightforward approach aims to mitigate bias by removing not only the sensitive variable but also variables that are correlated with it to reduce its indirect effects.\fcorr Several studies have highlighted the potential for non-sensitive variables to perpetuate the effects of sensitive variables in models, as discussed in \cite{lindholm2022discussion} and \cite{lindholm_2022b}.
Starting with explanatory variables $(X_{1},X_{2}, \dots, X_{p})$, the process involves the following steps:
    \begin{itemize}
    \item Measure the dependency between the sensitive variable, denoted as $S$, and each of the explanatory variables $X_{j}$.
    \item Based on the identified dependencies, create deletion scenarios, where each scenario represents a modeling instance with specific variables removed. Domain knowledge can also guide the selection of variables to delete.
    \item Build models for each scenario and assess their fairness and performance on predictions to identify the best models.
\end{itemize}
Additionally, scenarios can be automatically generated by setting a maximal dependency threshold and removing all variables with dependencies exceeding that threshold.



\subsubsection{Correlation remover}\label{sec:correlation}

Rather than deleting variables correlated with the sensitive variable, an alternative approach is to transform these variables to reduce bias while retaining some information. This method, suggested in studies like \cite{komiyama2017two} and \cite{frees2023discriminating}, involves removing information contained in $\boldsymbol{S}$ from $\boldsymbol{X}$ using regression models. The resulting residuals, denoted as $\boldsymbol{x}^\perp$, are used in place of $\boldsymbol{x}$, ensuring that $\boldsymbol{x}^\perp$ is orthogonal to the sensitive attributes $\boldsymbol{s}$. Formally, for a matrix of 
vectors $\boldsymbol{s}$, each variable $x_j$ is transformed as: $x^\perp_j = x_j - \alpha\boldsymbol{s}(\boldsymbol{s}^\top\boldsymbol{s})^{-1}\boldsymbol{s}^\top x_j$. $\alpha \in [0,1]$ is an hyperparameter that controls the level of correction applied on $x_j$.

It is important to note that the absence of correlation does not guarantee the absence of statistical dependence, especially in cases involving non-linear transformations of legitimate features. The bias mitigation achieved through this method is limited. However, for linear models like Generalized Linear Models (GLMs), this approach is consistent in breaking all the connections that models can establish.



\subsubsection{Fair-SMOTE adaptation}\label{sec:fair:smote}

The fair-SMOTE approach differs from previous techniques by introducing synthetic individuals into the dataset instead of altering the existing data. Its primary goal is to ensure equal gender representation, regardless of premium levels, potentially addressing under-representation issues of specific classes. It operates exclusively on the training dataset, leaving the test dataset untouched.

In a study by \cite{Chakraborty2021smote}, sampling methods directly on $Y$ were found to exacerbate bias as they do not consider the sensitive variable. They opted to sample based on $S\vert Y$ and $Y$. However, sampling on $Y$ in insurance pricing contexts may compromise unique target variable characteristics and decrease performance. Therefore, we avoid sampling on $Y$ and assess the ramifications in our use case.

To apply this method to continuous target variables, we discretize them, allowing delineation of resampling bins. The number of bins influences proximity to the continuous distribution and reduction of statistical bias from discretization. However, selecting too many bins may result in small sub-populations unfit for consistent simulations, requiring tuning based on target variable distributions. Once bins are designated, distributions of each $S$ modality are harmonized within each bin.

Instead of randomly selecting from the initial set, we introduce subtleties enabling distinct individuals' generation in specific scenarios. Two hyperparameters, the threshold $st$ and transformation factor $ft$, are defined and selected within the interval $[0,1]$. Using a $k$-nearest neighbor model, two closest individuals, $v_1$ and $v_2$, are identified from a randomly chosen individual $p$. New individuals' attributes are reconstructed column by column, preserving observed subgroup distributions. The pseudo code for this process is outlined in Algorithm \ref{algo:fair-smote} (Appendix \ref{secA1}). Continuous $Y$ values for newly generated individuals are calculated based on the algorithm's quantitative value rule.

Generative adversarial networks (GANs) could have been employed to simulate new individuals in a SMOTE approach or reconstruct $\boldsymbol{X}$ variables by generating similar individuals while minimizing dependence on $S$.



\subsection{In-processing} 

These approaches involve the inclusion of fairness constraints during the model calibration phase. The exponentiated gradient method, named after the game theory technique upon which it is founded, and its grid search version are both utilized.



\subsubsection{Exponentiated gradient}\label{sec:exponentiated}

\cite{agarwal2018exponen} present a bias mitigation approach that focuses on reducing bias within machine learning models. They begin by demonstrating that fairness definitions can be expressed as linear inequality sets of the following form: 
$$M\mu(m) \leq c,$$ 
In this representation, $M$ is a matrix, $c$ is a vector, and $\mu$ is a conditional moment vector. The vector $c$ provides the means to control the level at which each constraint is enforced by adjusting the values of $c_k$.

This formulation leads to an optimization problem within the context of statistical learning, defined as follows:
$$\min_{m\in\mathcal{M}} L(m) \text{ under the constraint that } M\mu(m) \leq c,$$ 
where $L$ represents a loss function employed for the evaluation of the models. We can observe that applying constraints to deterministic prediction functions can have a detrimental impact on performance. To mitigate this, they introduce the concept of "random functions," involving the use of a randomized predictor denoted as Q, which is drawn from $\Delta$, the set encompassing all distributions over $\mathcal{M}$, for the purpose of making predictions. In this approach, a predictor $m$, selected from $\mathcal{M}$, is sampled from $Q$ and subsequently used for prediction. Consequently, the prediction error is defined as $L(Q) = \sum_{m\in \mathcal{M}}Q(m)L(m)$ and the conditional moments are expressed as $\mu(Q) = \sum_{m\in\mathcal{M}}Q(m)\mu(m).$ This adaptation transforms the optimization problem into the following for
$$\min_{Q\in\Delta} L(Q) \text{ under the constraint that } M\mu(Q) \leq c.$$
To address this problem, we employ the "exponentiated gradient" algorithm, as recommended in \cite{freund1996gameth}. This game-theory-based approach pits the prediction function against the level of compliance with the constraint. The optimum is reached when any alteration of these elements results in a minimal loss of performance and an increase in constraint enforcement.

In theory, this approach has the potential to accommodate various fairness constraints in both binary and continuous cases. However, in practice, it has primarily been applied to the binary case. The implementation of equalized odds or demographic parity in the continuous case remains an unresolved challenge, and it is inconclusive whether this method is suitable for such cases.

A less stringent constraint known as "equality of the expected error" has been incorporated, specifically when $S$ is a discrete variable:
$$\text{ equality of } \mathbb{E}[\ell(Y,\widehat{Y})\vert S=s] \text{ across all } s.$$  This constraint aims to ensure that the model makes errors of similar magnitude, on average, regardless of the value of $S$, ultimately resulting in predictions of equal quality for different $s$ values. This concept can be reformulated as an inequality: 
$$\mathbb{E}[\ell(Y,\widehat{Y})\vert S=s] < \zeta, \forall s.$$ 
In this formulation, the hyperparameter $\zeta$ controls the acceptable error margin beyond which the constraint may be violated. Additionally, the error from a prediction can be further constrained by introducing the hyperparameter $M$, leading to the inequality:
$$\mathbb{E}[\min(\{\ell(Y,\widehat{Y})\vert S=s\},M)] < \zeta, \forall s.$$ This approach allows the incorporation of a threshold to limit the extent of deviations considered, particularly when the algorithm encounters convergence challenges.



\subsubsection{Grid search approach}\label{sec:grid}

In their exploration of bias mitigation, \cite{agarwal2018exponen} highlight an intriguing possibility when the sensitive variable is binary. In this context, selecting the most suitable hyperparameters becomes crucial, transforming the pursuit of the optimal solution into a challenge of identifying two interconnected parameters through an equation.

However, in the continuous case, a grid search methodology becomes vital for uncovering suboptimal solutions to the inequality mentioned earlier. When the exponentiated gradient method fails to converge for a given value of $\zeta$ or computational time becomes prohibitive, grid search offers an alternative. It systematically explores specific or random regions within the solution space, identifying a solution within a predefined time frame.

\subsection{Post-processing}

This approach involves transforming model predictions to enhance fairness. For instance, logistic regression allows influencing the model's behavior with respect to different classes through computed probabilities and decision thresholds without recalibrating predictions.

Historically, post-modeling techniques mainly adjust decision boundaries while considering fairness definitions, but they're less applicable to the continuous case due to the absence of distinct decision boundaries. Recent suggestions, like using Wasserstein barycenter on scores and the associated transport procedure by \cite{charpentier2023mitigating}, aim to address this gap. However, defining advantageous and disadvantageous premiums is challenging, as it depends on policyholder characteristics and inherent risk. The "individual fair redistribution" approach was introduced within this context.



\subsubsection{Fair redistribution  }\label{sec:fair:redist}

The methodology based on optimal transport, particularly the Wasserstein barycenter concept as utilized by \cite{charpentier2023mitigating}, involves implementing monotone transformations on premiums within distinct sensitive groups while maintaining the relative orderings of premiums within each group. Here, we'll focus on a uniform premium adjustment within each sensitive group. This approach adapts the flip-test to define bias, dividing premiums into fair shares and biases. The redistributed bias is then allocated to individual data points iteratively with the aim of minimizing bias given the adjusted premiums. \dcorr In this subsection, we will refer to $Y$ as premium to grasp the intuition behind our approach. The same reasoning can be made with any other continuous pricing outcome.
\fcorr

We introduce $\varepsilon$ to represent a measure of "individual fairness bias", which is defined as the difference between the estimated outcome $\widehat{y}$ and the fair premium $\widetilde{y}$. This estimation is initially determined by means of the quantity $\Delta_{\mathcal{V}}$, presented earlier. More specifically, we define $\varepsilon_i$ as $\Delta_{\mathcal{V}}^{i}$, as per the adaptation of the flip-test, which is elaborated upon in Section \ref{sec:2:3:2}.
But here, instead of completely rectifying the disparity in the premium due to the initially quantified bias with the equation:
$$\widetilde{y} = \widehat{y} - \varepsilon,$$
we adopt a strategy of partial fairness correction. This is accomplished by employing the following expression:
$$\widetilde{y} = \widehat{y} -\frac{\varepsilon}{\eta},$$
where $\eta$ is an hyperparameter constrained within the interval $[1,\infty)$, identical within a given group. When $\eta$ takes on a very large value, it corresponds to minimal correction, while as $\eta$ approaches 1, the level of correction is greater. $\eta$ was introduced because we noticed that correcting directly $\widehat{y}$ with $\varepsilon$ didn't lead to an increase of the fairness level. Individuals that had greater premiums than their neighborhood (as defined by the adapted flip-test) now had smaller premiums and vice versa. By bringing an individual's premium closer to the average premium of his opposite-gender neighbors, it potentially moves away from the premiums of other individuals whose neighborhood it made up. The conclusions of our first works were that a too brutal correction wasn't efficient and that it will be necessary to correct both subgroups smoothly and simultaneously. Thus, we decided to introduce an alternative algorithm where for a given level of $\eta$, we correct slowly and iteratively the two subgroup's premiums while recalculating at each iteration the new bias level $\varepsilon$. After the final iteration, we obtain 

$$\widetilde{y}_{i} = \widehat{y}_{i} - \varepsilon_{i}^{final~bias},$$ where $\varepsilon_{i}^{final~bias}$ can be seen as the summation of all $\varepsilon_{i}$ applied to an individual during the iterative process. To control the level of correction and stop the algorithm when needed, we introduce a second hyperparameter $\zeta \in [0,\infty)$ that will be a threshold that defines the maximal level of acceptable correction. Given a gender s (s binary in our case), we define : 

$$\Sigma_{s} = \sum_{i=1}^{n_s} \varepsilon_{i} = \sum_{i=1}^{n_s}  \Delta_{\mathcal{V}}^{i},$$
the sum of the fairness biases of all $n_s$ individuals in a gender subgroup. When $\Sigma_{s}<\zeta$, the algorithm will be stopped. This way, $\zeta$ can be used to control the tradeoff between fairness and performance, larger corrections leading to better fairness but a degradation of the quality of the premiums. The quality of the premiums after redistribution is measured using two metrics : 

\begin{itemize}
    \item Redistribution integrity, measures to which extent the redistribution has altered the scope of the premium's distribution : $$\frac{(\max \widetilde{y} - \min \widetilde{y}) }{(\max \widehat{y}- \min \widehat{y}) }.$$ 
    \item Global variation. Within the framework of the optimal transport approach, the fair premium is characterized by a balance property, corresponding to a null global variation. Measures the lost in premium induced by the redistribution : $$\sum_{i=1}^{n} \widetilde{y}_i-  \sum_{i=1}^{n} \widehat{y}_i.$$ 
\end{itemize}

With these elements, we detail the alternative approach in the following lines.  
Initialize by choosing $\eta, \zeta$ and setting a subgroup of $s$ to start with $(s=0)$. At each iteration, the treated subgroup will be switched and the bias will be recalculated. While $\Sigma_s\geq\zeta$ repeat the following steps :   
\begin{enumerate}
 
    \item Correct the premiums for the $s$ gender subgroup (update $\widehat y_i$): $$\widehat{y}_{i} = \widehat{y}_{i} - \frac{\varepsilon_{i}}{\eta},~~\forall i \text{~with~} s_{i} = s.$$
    \item Switch gender subgroup, take the opposite gender :
    
    $$
\begin{cases}
    \text{if } s=0, \text{ then } s=1,\\
    \text{if } s=1, \text{ then } s=0.
\end{cases}$$

    \item Measure the bias $\varepsilon_i$ between individuals of gender $s$ and their neighbors of opposite gender using the flip-test method (update $\varepsilon_i$): 
    $$\varepsilon_i = \widehat{y}_{i} - \frac{1}{k} \sum_{x_{j}\in \mathcal{V}^s(\boldsymbol{x}_i)}\widehat{y}_{j} = \Delta_{\mathcal{V}}^{i},~~\forall i \text{~with~} s_{i} = s.$$
    
    \item Calculate the sum of the differences: 
    
    $$\Sigma_{s} = \sum_{i=1}^{n_s} \varepsilon_{i},$$ if this sum is greater than $\zeta$ then repeat these five steps else stop the iterative process. 
\end{enumerate} 

The genders are switched before steps 3 and 4 because, after correction on a given gender subgroup, we want to recalculate the biases on the opposite gender subgroup and perform the check on the new $\Sigma_s$. If $\Sigma_s$ is still above our threshold $\zeta$, then we will repeat step 1 before switching for the next verification. This is because each correction on $\widehat y$ leads to a new fairness state which must be reevaluated with $\varepsilon_i$ to ensure a proper correction in the next iteration. Performing the check on the opposite gender ensures that the algorithm doesn't stop immediately after one gender subgroup has obtained sufficient corrections. 
As hyperparameters, $\zeta$ and $\eta$ have to be optimized to ensure the best possible results with regards to the redistribution integrity and global variation metrics. 



\begin{table}[]
    \centering
    \resizebox{\textwidth}{!}{%
 \begin{tabular}{ | m{.15\textwidth}| m{.50\textwidth}| m{.50\textwidth} | } \cline{2-3}
  \multicolumn{1}{c|}{}  & \multicolumn{1}{c|}{pros} & \multicolumn{1}{c|}{cons} \\ \hline

   \rowcolor{lightgray} Pre-processing    & preserves the overall modeling process. Straightforward to implement and consume less computational time. &  can lead to substantial loss of information. Variables may be closely interrelated and might not allow for both bias reduction and information retention. Additionally, it's crucial to prevent the modeling process from introducing new biases related to fairness. \\\hline
     Total deletion   & allows simple variable selection. HGR coefficients facilitates the identification of dependencies that classical measures might overlook. &  depends on the ability of the remaining variables to compensate for the loss of information resulting from variable deletion.
 \\\hline
     Correlation remover   & achieves an intriguing balance between fairness and performance. It is relatively simple to implement, relying on linear regression. & requires quantitative explanatory variables and ideally a quantitative sensitive variable. After correction, discretization of the transformed variable’s distribution is necessary, along with the construction of a correspondence function between the original and transformed variables for predictions.\\ \hline
     Fair-SMOTE adaptation   & needs various scenarios and hyperparameters allow customization to address specific problem requirements while sampling. & seems to have a limited effect on historical fairness bias and interdependencies.\\ \hline
  
     \rowcolor{lightgray} In-processing    & leverages sensitive variable information to identify the optimal trade-off between performance and fairness. In theory, they are more likely to yield the best possible compromise. & can be challenging to implement and generalize. Even after successful implementation, there is no assurance of convergence, and computation times can become exponential. \\\hline   Exponentiated gradient & offers the advantage of providing a comprehensive framework for the direct integration of fairness constraints into machine learning models. &  may require exponential computation times, even when dealing with relatively simple constraints. Certain cost functions and the imposition of various constraints may necessitate fundamental restructuring of the system to accommodate the application of the exponentiated gradient method. \\\hline
     
     \rowcolor{lightgray} Post-processing    & avoids model recalibration, leading to shorter computation times, and produce outcomes less susceptible to bias contamination. &  relies on the quality of built models, premium, and bias models, to be effective. Applying mitigation using inaccurately estimated components can result in inconsistencies. \\\hline
     Fair redistribution    & addresses individual fairness within regression scenarios and can be tailored to suit the specific problem under consideration. & needs monitoring of the quality of the premium and KNN models. And, considering the current distribution constraints, an extra step will may be necessary building a grid that encompasses all premiums corrections or distributable model that predicts the corrected premiums. \\\hline

    \end{tabular}}
    \caption{Summary of pros and cons of the different mitigation approaches}
    \label{tab:pro_cons}
\end{table}

\section{Measuring biases in car insurance pricing}\label{sec:3}

\dcorr For this glass breakage guarantee pricing application, various business and operational constraints alongside statistical considerations are essential throughout the pricing process, such as distribution constraints. \fcorr Table \ref{tab:var_before} displays the variables extracted from a car insurance database, augmented with vehicle information. It includes variable names, descriptions, values, and statistics. Quantitative variables show mean and median values, while qualitative ones indicate the shares of the two most common categories.

\begin{table}[!ht]
\centering
\resizebox{\textwidth}{!}{%
\begin{tabular}{|cccc|}
\hline
Variable name & Description & Value taken & Statistics \\ \hline
\texttt{claim\_amount} & individual claims expenses (\text{\sffamily€}) & $[0,1825]$ & $\Bar{x} : 580\text{\sffamily€}~\vert~ Me : 596\text{\sffamily€}$\\[0.1cm]
\texttt{claim\_nb} & number of claims & $[0,5]$ & $0: 97.3\% ~\vert~ 1: 2.7\%$ \\[0.1cm]
\texttt{expo} & exposure by contract & $[0{.}34\%, 100\%]$ & $\Bar{x} : 79\% ~\vert~ Me : 95\%$ \\[0.1cm]
\texttt{year\_pol} & year the policy was purchased & $[2015,2020]$ & $2018: 18\%~\vert~ 2020: 17\%$ \\[0.1cm]
\texttt{driv\_age} & age of primary driver & $[18,77]$ & $\Bar{x}: 47 ~\vert~ Me: 45$ \\[0.1cm]
\texttt{driv\_yp} & number of years in portfolio & $[0, 12]$ & $\Bar{x} : 1 ~\vert~ Me : 1.98$ \\[0.1cm]
\texttt{area} & area code & 17 zones & $D: 29\% ~\vert~ F: 23\%$ \\[0.1cm]
\texttt{driv\_gender} & gender of primary driver & F, M & $M: 58{.}4\%$ \\[0.1cm]
\texttt{driv\_ly} &driver licence seniority & $[0, 44]$ & $\Bar{x} : 15 ~\vert~ Me : 17$ \\[0.1cm]
\texttt{driv\_2} & presence of secondary driver & 0, 1 & $0:69\%$ \\[0.1cm]
\texttt{veh\_age} & age in year of the vehicle & $[0,44]$ & $\Bar{x} : 5 ~\vert~ Me : 4$ \\[0.1cm]
\texttt{energy} & type of energy & 5 types & $D:57\%~\vert~ E:43\%$ \\[0.1cm]
\texttt{weight} & weight in kilograms & $[830, 3200]$ & $\Bar{x} : 1240 ~\vert~ Me : 1280$ \\[0.1cm]
\texttt{veh\_power} & vehicle power in KW & $[13, 220]$ & $\Bar{x} : 94 ~\vert~ Me : 91$ \\[0.1cm]
\texttt{veh\_price} & vehicule price & $[6.8k,65k]$ & $\Bar{x} : 21400\text{\sffamily€} ~\vert~ Me : 15425\text{\sffamily€}$ \\[0.1cm]
\texttt{box\_type} & type of gearbox & 2 types & $A : 91.1\%$ \\[0.1cm]
\texttt{claim\_hist} & claim occurrence in previous observed years & 0, 1 & $0:91.7\%$ \\ \hline
\end{tabular}}
\caption{Description of the dataset's variables.}
\label{tab:var_before}
\end{table}

The data underwent processing and analysis, including variable reprocessing and discretization, to construct a modeling database. \dcorr Output variables \texttt{frequency}, \texttt{average cost}, and \texttt{pure premium} were constructed using \texttt{claim\_amount}, \texttt{claim\_nb}, and \texttt{expo}. \fcorr Two additional variables were created: \texttt{zoning}, describing risk zones with ten categories, and \texttt{weight\_kw}, representing the weight of the vehicle divided by its power. Models were built using Generalized Linear Models (GLM), Random Forest, and Gradient Boosting, evaluated using Root Mean Square Error (RMSE) and Loss Ratios (LR), and optimized and validated using interpretability methods.

Following the pricing phase, GLM models were selected for retention due to their interpretability and ease of integration into production pricing tools. Despite the potential for black-box models to achieve better performance, the marginal gain did not justify their operational costs. However, calculations were performed on all models at each stage, and there were generally no significant deviations.

Furthermore, the pure premium model was favored over a combined cost and frequency model for efficiency according to defined metrics. Reference models were established by excluding gender. The study then delves into measuring and mitigating gender bias to refine the pricing process using developed tools.

To assess fairness bias, six different dependence measures are employed:
\begin{itemize}
    \item[1-2] \dcorr Kendall's tau and mean ratio : These provide an initial understanding of dependence, offering a comprehensive first impression,\fcorr 
    \item[3-4] Kolmogorov-Smirnov's test $p$-value and the JS divergence : These quantify dependence between distributions,
    \item[5] HGR (or HGR\_KDE) : A potent metric offering a nuanced perspective,
    \item[6] Flip-test adaptation : An individual-based fairness metric.
\end{itemize}

The first five measures focus on group fairness, specifically independence, disregarding variable $Y$. The sixth measure, the Flip-Test Adaptation, examines bias at the individual level, providing an alternative perspective.

\subsection{Bias on historical data}

Table \ref{tab:mes_before} displays the results of the dependence measures between each of the variables of interest and the sensitive variable.

\begin{table}[!ht]
\centering
\begin{tabular}{|ccccccc|}
\hline
$Y$ & Kendall & HGR\_KDE & KS ($p$-value) & Div\_JS & mean\_ratio & Flip-test \\ \hline
Average cost & -0.0796 & 0.0903 & 1.9426e-07 & 0.3217 & 1.1024 & -10.23\text{\sffamily€}\\
Frequency & -0.0197 & 0.3031 & 3.3333e-02 & 0.8413 & 1.2489 & -2.57\%\\
Premium & -0.0212 & 0.3106 & 3.8555e-02 & 0.8401 & 1.3450 & -7.88\text{\sffamily€}\\ \hline
\end{tabular}
\caption{Dependence between $Y$ and $S$ before modeling}
\label{tab:mes_before}
\end{table}

Kendall's tau indicates a weak dependence between the variables of interest ($Y$ and $S$), with a negative sign suggesting slightly weaker values for women. HGR also detects slightly stronger relationships, consistent with Kendall's tau. The KS test highlights significant differences between male and female distributions.

\dcorr According to the flip-test, compared to similar male policyholders, women have an average cost €10 lower. This observation aligns with the negative sign of Kendall's tau and the mean ratio values, indicating lower averages for women compared to men. \fcorr

\begin{figure}[!ht]
    \centering
    \includegraphics[width=0.7\textwidth]{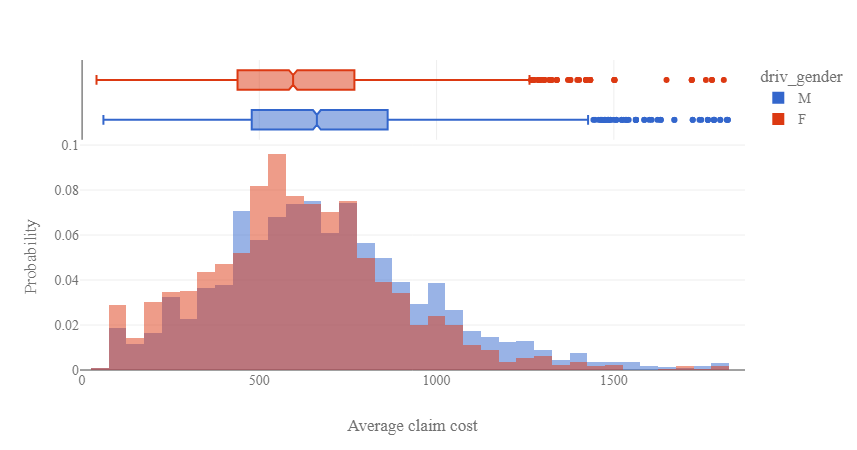}
    \caption{Historical average cost and premium distribution by gender.}
    \label{fig:distri_ac}
\end{figure}

\begin{figure}[!ht]
    \centering
    \includegraphics[width=0.7\textwidth]{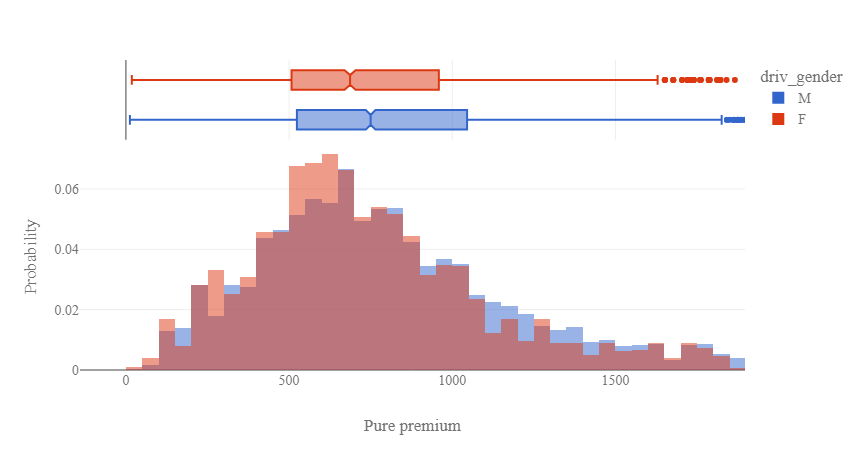}
    \caption{This histogram displays the historical premium distribution by gender. The premium is derived from the average cost times frequency distribution, with zeros excluded to facilitate the visualization of the remaining distribution.}
    \label{fig:distri_pp}
\end{figure}

Figures \ref{fig:distri_ac} and \ref{fig:distri_pp} analyze the distributions of these variables with respect to gender. For the frequency variable, distributions between $[0,1)$ for males and females are similar, representing 97.3\% of the population. At $1$, there's a 15\% higher representation of women compared to men, accounting for 1.2\% of the population. Above $1$, men are more strongly represented, with an 18\% higher presence compared to women, representing 1.5\% of the population.

The observed distributions suggest that, on average, men generally exhibit higher values for claim cost, frequency, and premium compared to women in historical data. However, interpreting these figures in absolute terms poses challenges due to the difficulty in establishing critical thresholds, given the lack of precedents in this type of problem. Nonetheless, these initial findings imply that gender has an impact on the variables of interest. Besides the class imbalance (more men than women), bias appears evident in the historical data. Such differences are well-documented in automobile pricing and are typically addressed by either excluding gender as a factor or rebalancing the model outputs. \dcorr It's worth noting that risk exposure may differ between gender classes due to various factors (see \cite{ayuso_2016}). To mitigate this effect, our dataset comprises individuals who subscribed to the same mileage package, providing the most accurate information available on vehicle usage.\fcorr

\subsection{Bias after modeling using gender}

By definition, the best-performing model would include gender as a feature because it leverages all the available information for risk modeling. However, it is also the most unfair model concerning gender since a clear distinction between men and women is directly visible in the resulting premiums.
To create this model, gender is reintroduced into the modeling process. Once constructed, predictions $\widehat{Y}$ are obtained, and the dependence between $S$ and $\widehat{Y}$ is measured. Table \ref{tab:mes_afteringenre} presents the results of these different measures.

\begin{table}[!ht]
\centering
\begin{tabular}{|ccccccc|}
\hline
$\widehat{Y}$ & Kendall & HGR\_KDE & KS ($p$-value) & Div\_JS & mean\_ratio & Flip-test \\ \hline
Average cost & -0.1824 & 0.2241 & 0.0000e+00 & 0.3825 & 1.0960 & -3.81\text{\sffamily€}\\
Frequency & -0.1949 & 0.3144 & 0.0000e+00 & 0.7333 & 1.2219 & -1.09\% \\
Premium & -0.2101 & 0.3268 & 0.0000e+00 & 0.7116 & 1.3524 & -1.44\text{\sffamily€} \\ \hline
\end{tabular}
\caption{Dependence between $\widehat{Y}$ and $S$ after modeling containing the gender}
\label{tab:mes_afteringenre}
\end{table}

\begin{figure}[!ht]
    \centering
    \includegraphics[width=0.7\textwidth]{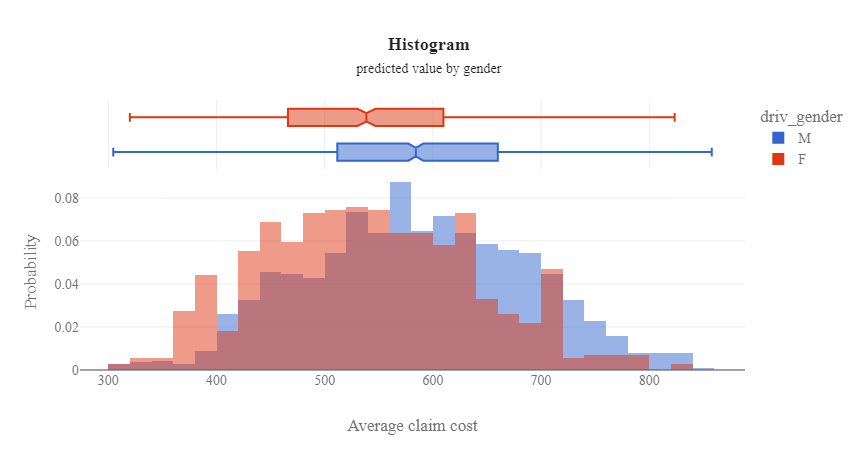}
    \caption{predicted average cost by gender.}
    \label{fig:distri_Ychap_cc}
\end{figure}

\begin{figure}[!ht]
    \centering
    \includegraphics[width=0.7\textwidth]{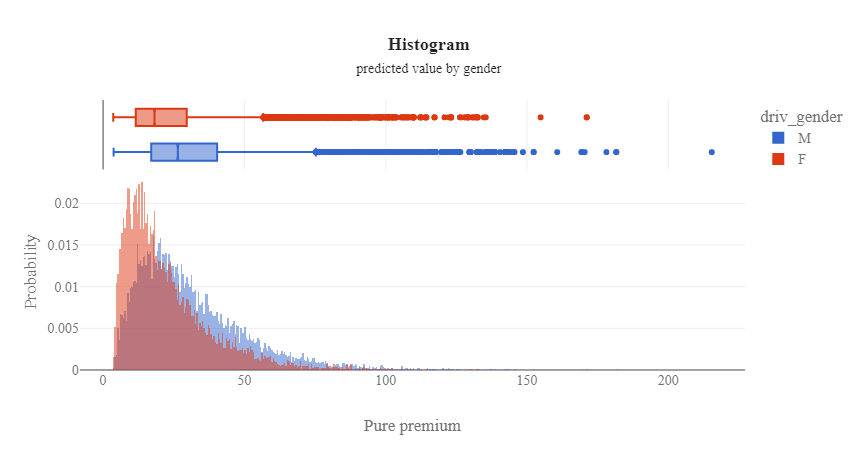}
    \caption{predicted pure premium by gender.}
    \label{fig:distri_Ychap_pp}
\end{figure}

\begin{figure}[!ht]
    \centering
    \includegraphics[width=0.7\textwidth]{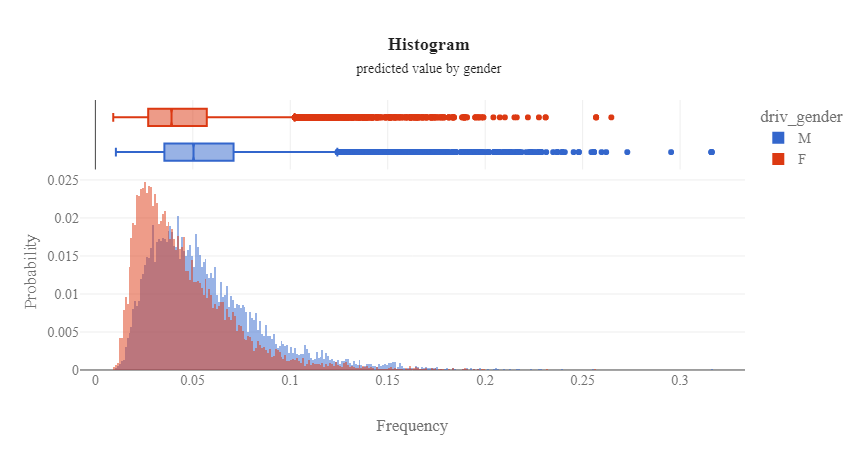}
    \caption{predicted frequency by gender.}
    \label{fig:distri_Ychap_fq}
\end{figure}

The dependencies between the interest variables and the sensitive variable were amplified according to almost all measures. Measures leveraging distributions, like div\_JS and the flip-test, suggest weaker dependencies because $\widehat{Y}$ has less dispersion than $Y$. Kendall's tau detects triple the dependence for the cost variable, implying a somewhat simpler detection of the dependency structure. Thus, the constructed models not only replicated the historical bias present in the data but also exacerbated it. Figures \ref{fig:distri_Ychap_cc}, \ref{fig:distri_Ychap_pp}, and \ref{fig:distri_Ychap_fq} analyze the distributions of these predicted interest variables concerning gender.


The mismatch between the distributions is more pronounced post-modeling, with larger portions of the distributions not overlapping. The differences are more evident compared to the historical data case.

How can we explain this bias amplification? The interpretation of the constructed models reveals that gender plays a significant role in the prediction processes. By examining coefficients and importance measures, it becomes evident that gender is among the most important variables in these models.

The greater measured bias in the constructed models compared to the historical data used for modeling can be explained by the interdependence of the variables in the dataset. There is a link between the explanatory variables selected and the variables of interest because they help understand the associated risk. However, these variables, besides their predictive abilities, may also be related to each other. For example, a vehicle's power may be linked to its price. These interdependencies pose challenges for fairness implementation because the so-called sensitive variables often have a notable influence on the distributions of the other observed variables. Thus, even if the observable interdependencies seem weak at first glance, their accumulation can magnify the role of the sensitive variable in the models.

\begin{figure}[!ht]
    \centering
    \includegraphics[width=\textwidth]{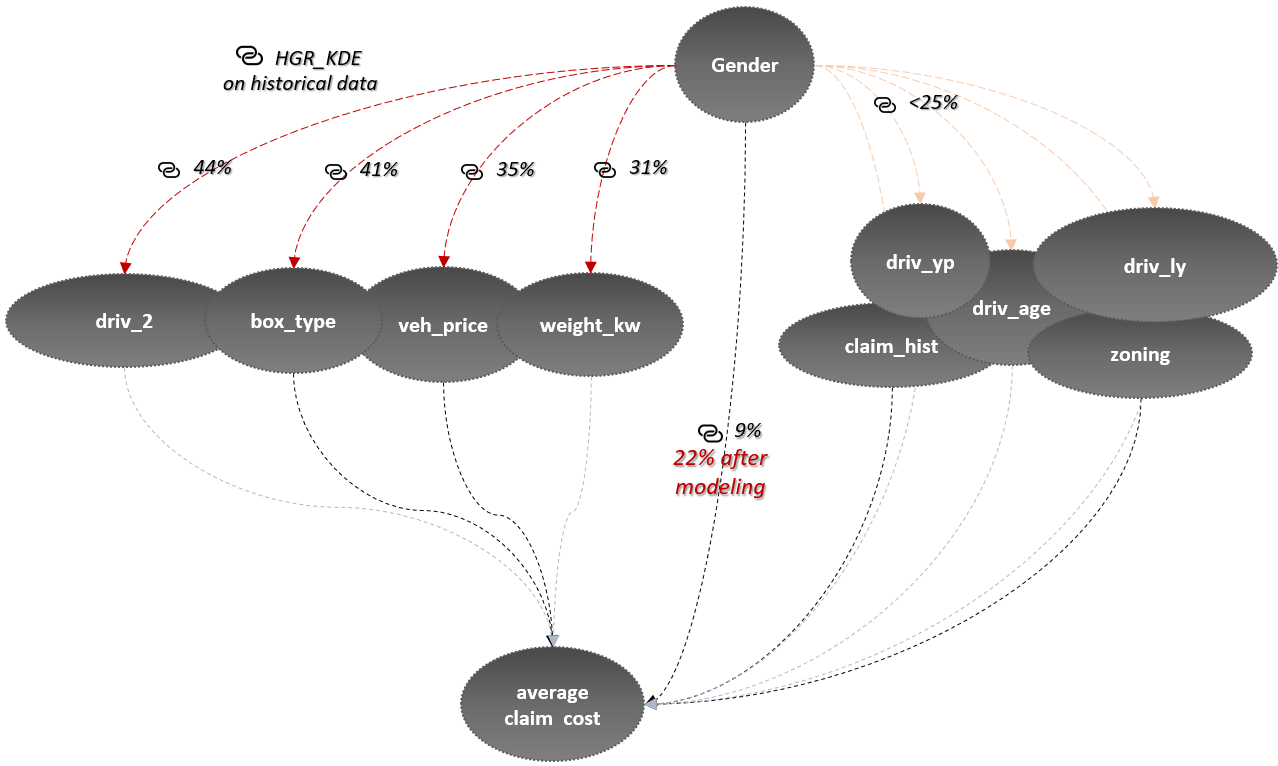}
    \caption{Direct and indirect effects of gender on the predicted average cost. The darker the line, the stronger the relationship.}
    \label{fig:causal_genre_aftergender}
\end{figure}

In the data used, gender is significantly linked with variables such as horsepower, weight, gearbox type, and vehicle price, in addition to its minor relationships with other explanatory variables. Figure \ref{fig:causal_genre_aftergender} visually represents how gender directly influences the variables of interest and has a significant indirect effect through its relationships with other variables.

For instance, in the cost model, gender is ranked as the fourth most influential variable. Nevertheless, it's noteworthy that the most substantial variables within the model are those closely related to gender. Generally, sensitive variables exert a significant impact on individuals' behavior. For instance, age can influence risk-taking propensity, the level of responsibility, personal interests, and more. Similarly, ethnicity may correlate with factors such as geographical location, purchased services, and cultural preferences.

These analyses inevitably lead to a central question: What are the implications when gender is omitted or excluded from the modeling process? This raises the possibility that the observed interdependencies might have been overemphasized, and that the models have unduly relied on gender as a driving factor.

\subsection{Bias after modeling without gender}

By definition, a model that excludes gender as a variable aims to eliminate direct discrimination, ensuring that final outcomes are not discerned based on gender. However, it's acknowledged that this strategy doesn't address indirect discrimination as it overlooks the interrelationships between other explanatory variables and gender. Table \ref{tab:mes_afteroutgenre} presents findings on the association between $\widehat{Y}$ and gender in models where gender isn't included as an explanatory variable.

\begin{table}[!ht]
\centering
\begin{tabular}{|ccccccc|}
\hline
$\widehat{Y}$ & Kendall & HGR\_KDE & KS ($p$-value) & Div\_JS & mean\_ratio & Flip-test \\\hline
Average cost & -0.1681 & 0.2111 & 9.7460e-39 & 0.3288 & 1.0118 & -2.71\text{\sffamily€}\\
Frequency & -0.1309 & 0.2741 & 0.0000e+00 & 0.6493 & 1.1573 & -1.07\%\\
Premium & -0.1733 & 0.2897 & 0.0000e+00 & 0.7226 & 1.2694 & -0.88\text{\sffamily€}\\ \hline
\end{tabular}
\caption{Dependence between $\widehat{Y}$ and $S$ after modeling without gender}
\label{tab:mes_afteroutgenre}
\end{table}

The results show a minimal reduction in bias compared to models where gender is included, indicating that gender's influence remains pronounced even when it's absent from the models. While it may seem counterintuitive that premiums no longer differentiate by gender, there are still distinctions observed based on gender, primarily due to indirect interdependencies within the data. For example, varying premiums based on the presence of secondary drivers indirectly leads to varied premiums for each gender because secondary drivers have different distributions by gender. In complex multivariate models, such distinctions based on gender can arise from the combination of multiple variables. For instance, a combination of factors such as vehicle power, gearbox type, and zoning may result in significant gender imbalances within the dataset. When examining the intersection of zoning and vehicle prices, notable gender imbalances become evident in the data. For instance, 82\% of individuals residing in the 10th zone and owning a vehicle costing over €30,000 are male. Consequently, any substantial premium difference within this segment compared to others significantly impacts men, leading to gender-based premium disparities, despite the absence of explicit gender classification.

\subsection{A different modeling approach}

These findings highlight that implementing fairness through omission does not offer a comprehensive solution to discrimination. An alternative modeling approach is employed to assess dependency levels. Instead of excluding the gender variable, it is included in the model, and then the outputs are post-processed to ensure gender-insensitive pricing, aligning with strategies used by insurers to comply with regulations like the gender directive. The specific approach calculates a weighted average of male and female model outputs for all segments, but the results show similar dependencies between the new values of $\widehat{Y}$ and $S$ as previous modeling strategies. This method fails to resolve interdependencies, as reweighting the outputs perpetuates existing imbalances.

Consequently, gender's influence persists within the modeling process, even after its omission or conventional adjustment. Some scholars refer to this as the "reconstruction" of the sensitive variable, where bias continues to manifest due to interdependencies between explanatory variables and the sensitive variable. This study exemplifies such a scenario.

In summary, bias exists in historical data, and regardless of the modeling approach applied, it is perpetuated, primarily due to interdependencies. The subsequent section explores bias mitigation while maintaining acceptable performance, using the model constructed without gender as reference.

\section{Bias mitigation in car insurance pricing}\label{sec:5}

The models have been constructed (using standard machine learning approaches), performance and biases have been assessed, and the next step involves mitigating bias while upholding the performance level. This section provides the implementation specifics and outcomes of the various bias mitigation methods introduced in Section \ref{sec:bias_mitigation}. 

\subsection{Total deletion}

\begin{figure}[!ht]
    \centering
    \includegraphics[width=\textwidth]{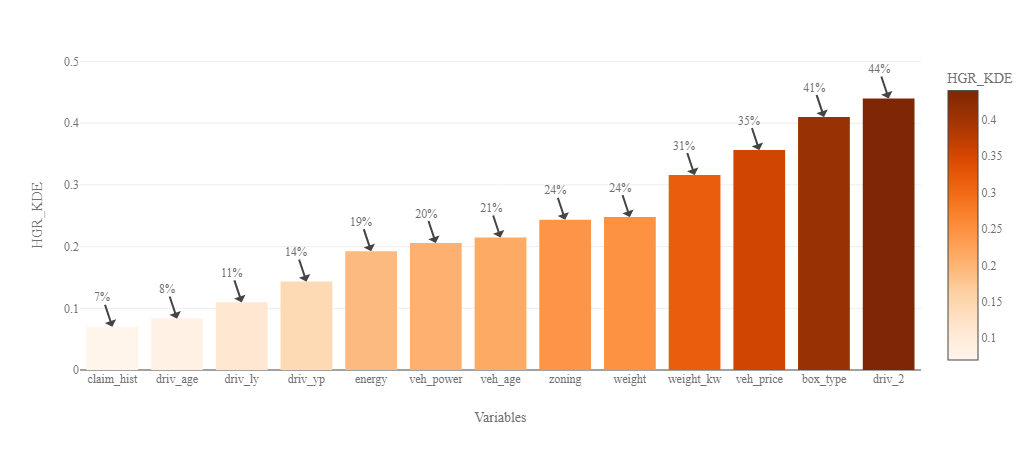}
    \caption{Interdependence between gender and other explanatory variables}
    \label{fig:dependancesetx}
\end{figure}

To establish the deletion scenarios, we analyze the relationship between sensitive and non-sensitive variables. The results are depicted in Figure \ref{fig:dependancesetx}. 

\begin{figure}[!ht]
    \centering
    \includegraphics[width=0.8\textwidth]{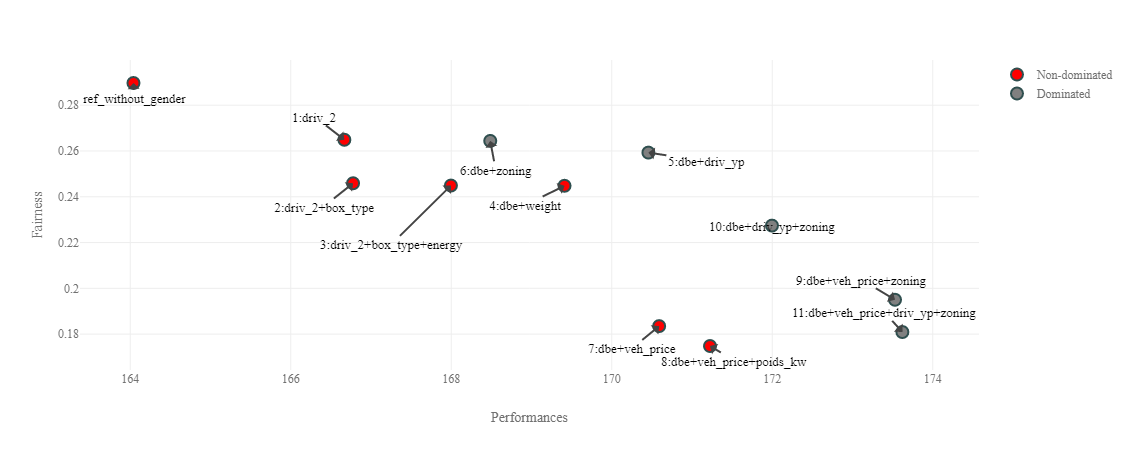}
    \caption{model's fairness according to their performance}
    \label{fig:combat_sup}
\end{figure}

By examining these dependency levels, $\alpha$ can assume values within the range of $(8\%,41\%].$ Scenarios are developed by iteratively adjusting the threshold, considering the relationships between the variables. The chosen scenarios are as follows:
\begin{itemize}
    \item Scenario 1: \texttt{driv\_2};
    \item Scenario 2: \texttt{driv\_2} and \texttt{box\_type};
    \item Scenario 3: \texttt{driv\_2}, \texttt{box\_type} and \texttt{energy};
    \item Scenario 4: \texttt{box\_type}, \texttt{driv\_2}, \texttt{energy} and \texttt{weight};
    \item Scenario 5: \texttt{box\_type}, \texttt{driv\_2}, \texttt{energy} and \texttt{anc\_cp};
    \item Scenario 6: \texttt{box\_type}, \texttt{driv\_2}, \texttt{energy}, and \texttt{zoning};
    \item Scenario 7: \texttt{box\_type}, \texttt{driv\_2}, \texttt{energy} and \texttt{veh\_price};
    \item Scenario 8: \texttt{box\_type}, \texttt{driv\_2}, \texttt{energy}, \texttt{veh\_price} and \texttt{weight\_kw};
    \item Scenario 9: \texttt{box\_type}, \texttt{driv\_2}, \texttt{energy}, \texttt{veh\_price} and \texttt{zoning};
    \item Scenario 10: \texttt{box\_type}, \texttt{driv\_2}, \texttt{energy}, \texttt{weight} and \texttt{zoning};
    \item Scenario 11: \texttt{box\_type}, \texttt{driv\_2}, \texttt{energy}, \texttt{veh\_price}, \texttt{weight\_kw} and \texttt{zoning}.
\end{itemize} 
The gender variable is eliminated from all models. Figure \ref{fig:combat_sup} illustrates the fairness and performance levels of each scenario. The non-dominated scenarios are highlighted in red. The abbreviation \texttt{dbe} refers to the variables \texttt{driv\_2}, \texttt{box\_type} and \texttt{energy}. For example, scenario 6 is represented on the graph as \texttt{dbe+zoning}.

Among the non-dominated scenarios, a choice must be made based on the desired trade-off as determined by decision-makers. For instance, the objective might be to attain the fairest model while tolerating a maximum performance loss of 5\%. Scenarios 8, 9, 10, and 11, although among the fairest, would no longer meet this criterion. Upon analyzing the graph, the 7th scenario, labeled as \texttt{dbe+veh\_price}, emerges as an interesting trade-off option, as it significantly reduces bias, almost halving it, with an acceptable performance loss. Compared to the reference model, a bias reduction of 37\% is achieved in exchange for a 3.9\% drop in performance. The 7th scenario model is further examined.

It appears that the model has learned to rely more on other variables to compensate for the absence of the \texttt{veh\_price} variable. For instance, some variables like \texttt{zoning}, \texttt{driv\_ly}, and \texttt{veh\_age} now have larger coefficients and play a more significant role in the prediction process. However, the performance loss still indicates that these omitted variables cannot be entirely replaced optimally. Furthermore, this new model leads to a slightly degraded equilibrium in pricing compared to the reference equilibrium. The LR metric obtained is 99.3\% compared to the reference LR of 99.7\%. It is essential to note that once the input data are modified, a reevaluation and validation of the model are necessary.

Hence, this initial approach can yield satisfactory results depending on the quality of the explanatory variables and the acceptable level of performance loss. It is a straightforward method to implement and underscores the importance of addressing fairness concerns from the initial stages of data processing and variable selection.

\subsection{Correlation remover}

To apply the methodology introduced in Section \ref{sec:correlation}, we have chosen to retain qualitative variables while focusing solely on transforming quantitative variables. The quantitative variables subject to transformation include: \texttt{veh\_price}, \texttt{weight\_kw}, \texttt{veh\_age}, \texttt{driv\_yp}, and \texttt{driv\_ly}. We uniformly select $100$ values of $\alpha$ within the interval $[0,1]$, including both endpoints. For each $\alpha$ value, we correct our quantitative variables, build our model, and measure performance and fairness.

From these results, two observations arise. First, there is no clear trend between the fairness level and the hyperparameter $\alpha$. Typically, one might expect fairness and performance to decrease as $\alpha$ increases, but this trend is not evident. The relationship between fairness, performance levels, and $\alpha$ appears random, lacking a discernible pattern. Second, using this method, some models outperform the reference model. For example, the model with $\alpha=85\%$ shows lower bias and slightly better performance. While the performance gain is only around 0.5\%, it comes with an 11\% improvement in fairness. However, the efficacy of this method relies on the linear model's ability to detect relationships, ensuring that residuals are unbiased. Nonetheless, examining linear correlations reveals that the highest dependence between $S$ and other explanatory variables is 8\%, significantly limiting suppression capacity in this specific case.

While this method seems promising, it presents several limitations within the insurance pricing context. First, in pricing models, it is more common to use discrete or discretized variables as inputs. Ideally, continuous variables should be discretized after suppression to ensure interpretability. However, as the values represent residuals, they may lose interpretability. Additionally, in practical production scenarios, transforming customer attributes into predicted premiums may be complicated by this method. Lastly, the non-intuitive model behavior in response to changes in $\alpha$ lacks a clear explanation. Considering these factors, while a model outperforming the reference model was achieved, adopting it is challenging due to associated limitations. This method may find greater success when applied to a continuous sensitive variable.

\subsection{Fair-SMOTE adaptation}

Balancing the number of bins and the size of each subpopulation, we have opted for a segmentation into seven bins, with the following breakdown:
\begin{itemize}
 \item bin 1 if $y=0$;
 \item bin 2 if $y \in (0,250]$;
 \item bin 3 if $y \in (250,500]$;
 \item bin 4 if $y \in (500,750]$;
 \item bin 5 if $y \in (750,1000]$;
 \item bin 6 if $y \in (1000,1500]$;
 \item bin 7 if $y \in (1500,+\infty)$.
\end{itemize}
~~\\

The number of bins serves as a hyperparameter, resulting in various binning scenarios. The selection of these bins depends on their alignment with the observed trends in the premium distribution.

Following the bin selection, the next step is to determine the scope of resampling. As previously discussed, resampling will only be performed on $S\vert Y$ since resampling on $Y$ would alter the specific distribution needed for the GLM. Although for algorithms like random forest, which make no distribution assumptions, resampling on $Y$ was attempted, it led to a notably lower LR due to the high premium levels predicted by the models.

\begin{figure}[!ht]
    \centering
    \includegraphics[width=.9\textwidth]{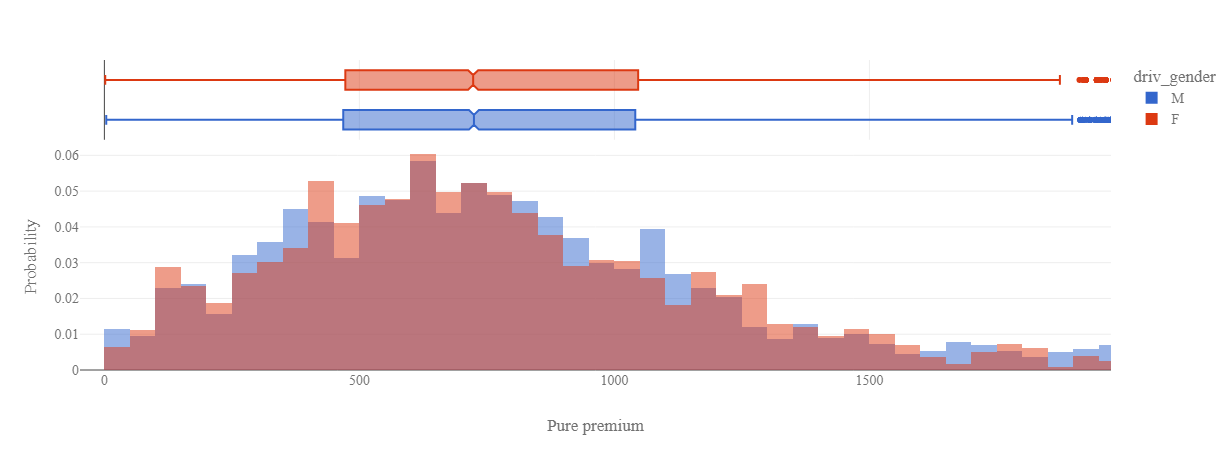}
    \caption{Distribution of $Y$ after resampling.}
    \label{fig:distrechanrechan}
\end{figure}

It's worth noting that the creators of the traditional fair-SMOTE method recommend setting $st$ and $ft$ to 0.8 each. This configuration is retained for consistency. The distribution after resampling is illustrated in Figure \ref{fig:distrechanrechan}.
 
In total, 9588 individuals were generated, constituting a substantial expansion of the learning dataset by 16.1\%. The resulting distribution of this enlarged dataset remains in alignment with the initial observations, and the distinctions between genders are notably less prominent. The box plots demonstrate a substantial overlap, with the noticeable shift observed in the historical data significantly mitigated. Therefore, this resampling procedure appears to be effective. Following the preparation of the augmented training data, the models are reconstructed, and the outcomes are presented in Table \ref{tab:resfairsmote}.

\begin{table}[!ht]
\centering
\begin{tabular}{|cccc|}
\hline
Models & RMSE & HGR\_KDE & LR \\ \hline
Reference model & 164.21 & 28.97\% & 99.71\% \\
Model after fair-SMOTE & 164.88 & 28.05\% & 99.69\% \\ \hline
\end{tabular}
\caption{Comparative table of results after fair-SMOTE.}
\label{tab:resfairsmote}
\end{table}

The initial distribution shows reduced skewness, but the outcomes on the test dataset still exhibit skewed results. The improvement in fairness is minimal, with less than a 4\% enhancement. Performance metrics and the LR remain similar to the baseline. Therefore, rebalancing the distributions solely based on $S\vert Y$ seems inadequate for achieving fairness. This could be because the bias needing mitigation in car pricing does not stem from a representation issue concerning $Y$, but rather from historical differences in male and female premiums. Thus, while resampling ensures both genders receive sufficient data for equitable treatment during training, it doesn't address the interdependencies between variables, crucial for ensuring fairness.

Sensitivity analyses on various modeling choices were conducted to explore whether these choices contributed to the ineffectiveness of our approach. Changes to hyperparameters $ft$ and $st$ didn't result in improved outcomes. Adjusting the resampling bins also didn't lead to significant enhancements. Increasing the number of bins introduced more irregularities in the distribution of $Y$ and reduced model performance.

A significant departure from conventional practices in the literature was limiting the resampling process solely to $S \vert Y$ rather than conducting full resampling on $Y$. Full resampling of $Y$ altered its distribution drastically, generating an excessive number of artificial instances and failing to produce satisfactory results. To explore the potential benefits of resampling on $Y$ while preserving the overall distribution shape, partial rebalancing of $Y$ values was considered. For instance, in scenarios where $Y$ has only two distinct values, such as $\{25, 35\}$, Tables \ref{tab:ex1} to \ref{tab:ex4} demonstrate the various forms of resampling undertaken.

\begin{table}[!ht]
\parbox{0.45\linewidth}{
\centering
\begin{tabular}{|cccc|}
\hline
Y/S & F & H & Total \\ \hline
25 \text{\sffamily€} & 250 & 150 & 400 \\
35 \text{\sffamily€} & 120 & 170 & 290 \\
Total & 370 & 320 & 690 \\ \hline
\end{tabular}
\caption{initial distribution}
 \label{tab:ex1}
}
\hfill
\parbox{0.45\linewidth}{
\centering
\begin{tabular}{|cccc|}
\hline
Y/S & F & H & Total \\ \hline
25 \text{\sffamily€} & 250 & 250 & 500 \\
35 \text{\sffamily€} & 170 & 170 & 340 \\
Total & 420 & 420 & 840 \\ \hline
\end{tabular}
\caption{Balance on $S\vert Y$.}
 \label{tab:ex2}
}
\hfill
\parbox{0.45\linewidth}{
\centering
\begin{tabular}{|cccc|}
\hline
Y/S & F & H & Total \\ \hline
25 \text{\sffamily€} & 250 & 250 & 500 \\
35 \text{\sffamily€} & 250 & 250 & 500 \\
Total & 500 & 500 & 1000 \\ \hline
\end{tabular}
\caption{Balance on $S\vert Y $ and $Y$.}
 \label{tab:ex3}
}
\hfill
\parbox{0.55\linewidth}{
\centering
\begin{tabular}{|cccc|}
\hline
Y/S & F & H & Total \\ \hline
25 \text{\sffamily€} & 250 & 250 & 500 \\
35 \text{\sffamily€} & 200 & 200 & 400 \\
Total & 450 & 450 & 900 \\ \hline
\end{tabular}
\caption{Balance on $S\vert Y$ and partially on $Y$.}
 \label{tab:ex4}
}
\end{table}

Multiple scenarios involving partial resampling of $Y$ were explored. The results indicate that as the number of artificially generated individuals increases, model performance worsens without simultaneous improvements in fairness. For example, expanding the population by 20\% through resampling led to an increased RMSE of 177, while the HGR fairness metric remained at 28.05. Ultimately, these approaches failed to produce better results.

An unexplored area for potential improvement involves considering the explanatory variables during the resampling process. Instead of resampling solely to achieve equity based on $S\vert Y$, an alternative approach is resampling based on $S\vert X_1, \dots, X_p, Y$. The selection of variables for conditioning is crucial here, as they can potentially amplify existing biases. Implementing this approach is complex, as it requires overseeing the quality of resampling on individual variables and their interactions while maintaining the overall consistency of premium distributions. Although this approach hasn't been extensively explored in this study, integrating more efficient methods for addressing interdependencies between variables may be considered in future research.

In summary, the various pre-modeling bias mitigation strategies have provided insights into the data's structure and identified elements with significant influence on bias levels. They offer valuable tools for addressing ethical concerns during the data preprocessing phase.

\subsection{Exponentiated gradient}

To achieve the model with the lowest permissible error level, bounded by $\zeta$, a specific approach is adopted. The method starts with the lowest feasible $\zeta$ value, which is initially set at $10^{-5}$, and progressively increases it in multiples of 10 until the algorithm converges. The first convergence occurred when $\zeta = 10^{-1}$. Once the method converges to the optimal solution, the grid search and the $M$ parameter become unnecessary.

However, for $\zeta<10^{-1}$, these tools can help identify suboptimal solutions that may outperform the solution found with larger tolerances. During testing, values of $M$ ranging from 0.5 to 500 were employed with grid sizes of 3000. The results obtained were not as favorable as those achieved with convergence. It is essential to note that the grids explored were relatively small compared to the dimensionality of the dataset. Given more computational power and time, improved results might potentially be attainable.

The outcomes for the best model are presented in Table \ref{tab:resexpono}.

\begin{table}[!ht]
\centering
\begin{tabular}{|cccc|}
\hline
Models & RMSE & HGR\_KDE & LR \\ \hline
Reference model & 164.21 & 28.97\% & 99.71\% \\
Model after mitigation & 164.30 & 31.74\% & 99.70\% \\ \hline
\end{tabular}
\caption{Evaluation of the results obtained after application of the exponentiated gradient.}
\label{tab:resexpono}
\end{table}

The model obtained after mitigation exhibits reduced efficiency and increased unfairness compared to the reference model. While the loss of performance is negligible, the model experiences a 9\% higher HGR after the application of the mitigation method. These outcomes can be attributed to the fact that the error constraint implemented does not enforce any form of independence between $\widehat Y$ and $S$. In other words, the error rate per gender can be the same without ensuring fair treatment of the genders.

In addition to the inability to accommodate alternative fairness definitions, this method presents exponential computation times. The grid search approach and the parameter $M$ offer an alternative but lead to suboptimal results. In the rapidly evolving domain of bias mitigation, the aspect of mitigation during modeling stands as one of the most challenging and least developed. The limited availability of methods primarily focused on binary classification, combined with the high customization required for their adaptation, compounds the challenge. Nonetheless, the method explored here, implemented under a relatively straightforward fairness constraint, serves to illustrate the limitations associated with implementing fairness as an optimization constraint. Furthermore, the quest for accessible, generalizable, and stable methods remains a significant mathematical challenge that must consider the constraints inherent to the field of pricing.

\subsection{Fair redistribution}

This method employs the flip test's adaptation to measure individual fairness. We optimized the $k$-nearest neighbors used in the flip test through hyperparameter tuning. The objective was to select the most suitable variables that would minimize the differences between individuals in a test database. As discussed earlier, the goal is to develop a model that reduces the distance and bias between individuals.

\begin{figure}[!ht]
    \centering
    \includegraphics[width=.9\textwidth]{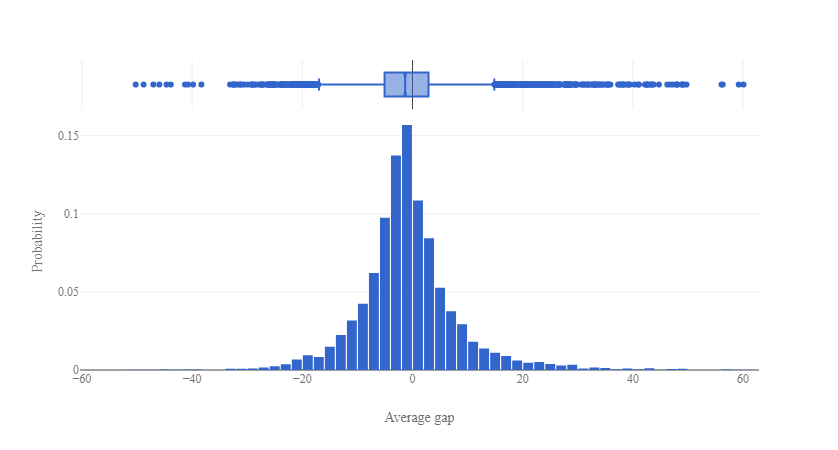}
    \caption{Histogram of $\varepsilon_i$ (gaps) between women and their male neighbors}
    \label{fig:hist_femme_flipt}
\end{figure}

The selected variables include: \texttt{driv\_yp}, \texttt{driv\_2}, \texttt{veh\_age}, \texttt{veh\_price}, \texttt{box\_type}, and \texttt{zoning}, with a number of neighbors set to 5 and a Manhattan distance metric ($\ell_1$). In Python, there is an attribute in the $k$-nearest neighbor algorithm that can automatically select the optimal neighbor search method based on factors like dimensionality, the number of individuals, and the data structure (e.g., whether the matrix is sparse or not). This automated method yielded the most favorable results. Figure \ref{fig:hist_femme_flipt} illustrates the distribution of disparities between women and their corresponding male neighbors.

Subsequently, for each individual in the database, the average disparity with their closest neighbors of the opposite gender is computed. These average discrepancies are incorporated into the original modeling dataset. An experiment is then conducted on a test dataset, encompassing a total cost of 359584\text{\sffamily€} and predicted premiums of 360884\text{\sffamily€} for 14000 observations. The Loss Ratio (LR) of 99.64\% aligns well with the overall LR of 99.7\%. On average, women's premiums are 0.8\text{\sffamily€} lower than those of the nearest men. Table \ref{tab:des_base_redis} provides a summary of the interesting aggregates calculated on the test dataset.

\begin{table}[!ht]
\centering
\begin{tabular}{|cccc|}
\hline
 Aggregates & Female & Male & Sum \\ \hline
Total expenses & 154880\text{\sffamily€} & 204704\text{\sffamily€} & 359584\text{\sffamily€} \\[0.1cm]
Predicted Premiums & 154953\text{\sffamily€} & 206931\text{\sffamily€} & 360884\text{\sffamily€} \\[0.1cm]
Exposure & 4776.6 & 7114.6 & 11891.2 \\[0.1cm]
Number of individuals & 6069 & 7931 & 14000 \\[0.1cm]
Average $\varepsilon$ & -0.8\text{\sffamily€} & 1.2\text{\sffamily€} & 0.4\text{\sffamily€} \\[0.1cm]
Sum of bias ($\Sigma_s$) & -17448\text{\sffamily€} & 22029\text{\sffamily€} & 4581\text{\sffamily€} \\ \hline
\end{tabular}
\caption{Some relevant aggregates on the test perimeter.}
\label{tab:des_base_redis}
\end{table}

The sum of individual fairness bias for women is 17448\text{\sffamily€}, while men have a sum of 22029\text{\sffamily€}. These disparities can be interpreted as follows: on average, women pay 17448\text{\sffamily€} less than men with similar characteristics. The signs of the differences align with expectations, reflecting that women pose lower risks compared to men. However, the differences do not fully offset each other, resulting in a residual difference of 4581\text{\sffamily€} (-17448\text{\sffamily€} + 22029\text{\sffamily€}). This remaining variance can be attributed, in part, to the larger representation of men in the database and the imperfections within the employed models.

To assess the influence of different values for $\zeta$ and $\eta$, and to identify the optimal combination, the redistribution process is repeated across a grid of values. This grid covers all possible combinations of $(\eta, \zeta)$, where $\eta$ takes on values from the set $\{2, 3, 4, 5, 6, 7, 8, 9, 10\}$ and $\zeta$ from the set $\{2500, 2000, 1500, 1000, 500, 100, 10, 1, 0.1\}$. Three factors are examined: computational time, global variation, and redistribution integrity.

When $\zeta\leq 100$, the redistribution process results in a substantial reduction in the range of the $\widehat Y$ distribution. This occurs because the correction needed to achieve such low levels of differences between subgroups is extensive, causing premiums to become increasingly clustered until they eventually converge towards the sample's mean premium. Consequently, pushing the redistribution method towards maximum convergence leads to an equilibrium where all individuals are assigned the same premium. While this equilibrium represents trivial fairness, it severely deteriorates the model's performance.
Therefore, it becomes imperative to guide the method towards optima where global variation is minimized while preserving the distribution of $\widehat Y$. Among the 81 tested combinations, only 10 of them emerged as non-dominated solutions based on criteria of redistribution integrity and global variation. Out of these 10, 5 had a redistribution integrity of less than 25\%, making them unacceptable despite having the smallest global variations. The two most promising scenarios are as follows:

\begin{enumerate}
    \item $\eta = 6$ and $\zeta = 2000$ for a global variation of $873$\text{\sffamily€} and an integrity of $88\%$;
    \item $\eta = 5$ and $\zeta = 2500$ for a global variation of $771$\text{\sffamily€} and an integrity of $86\%$.
\end{enumerate}
~~\\

\begin{figure}[!ht]
    \centering
    \includegraphics[width=.9\textwidth]{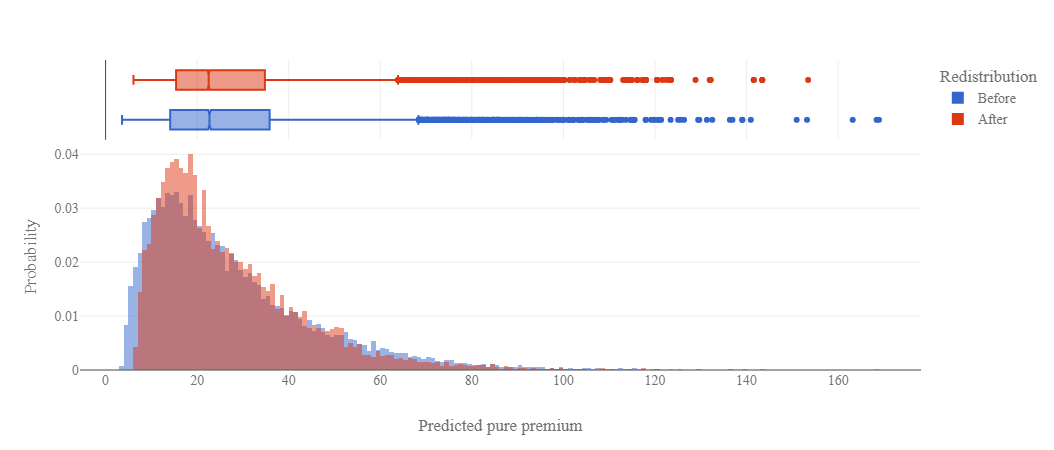}
    \caption{Distribution of $\widehat Y$ before and after redistribution}
    \label{fig:redisdistr}
\end{figure}

The scenario with 88\% fidelity is preferred, resulting in a sum of gaps of -603\text{\sffamily€} for women and +1682\text{\sffamily€} for men. Consequently, the total gap is reduced to 1079\text{\sffamily€}, representing a 76\% decrease compared to the initial gap of 4581€. The average bias are now 0.071\text{\sffamily€} for women and 0.14\text{\sffamily€} for men, all while preserving a distribution of $\widehat Y$ that remains faithful to the one prior to redistribution. Figure \ref{fig:redisdistr} illustrates the overlap of histograms before and after the redistribution.

Regarding performance, the premiums post-redistribution exhibit a Root Mean Square Error (RMSE) of 164.72, which closely aligns with the baseline value of 164.21. Despite the increase of 873€ in total premiums (global variation), the Loss Ratio (LR) only experiences a minor reduction from 99.64\% to 99.39\%. Thus, premiums maintain their consistency and calibration while effectively reducing the gaps between men and women. For values of $\zeta$ less than or equal to 100, most computation times are under five minutes, rendering this criterion less relevant for defining the best redistribution strategies.

The introduction of bias mitigation methods has prompted the examination of fairness considerations throughout various phases of the pricing process. The successful integration of these mitigations necessitates alignment with the diverse requirements and limitations inherent to pricing. Consequently, they should be operationally feasible while striving to maintain the performance of the models. Among the implemented methods, variable suppression, due to its simplicity, and redistribution, owing to its personalized approach, appear to yield the most promising outcomes. The other methods contribute to a more comprehensive understanding of the analyzed bias and may yield superior results in alternative datasets. An overview of the results stemming from the various mitigation strategies is presented in Table \ref{tab:recap_mitigation_app}.

\begin{table}[!ht]
\centering
\resizebox{\textwidth}{!}{%
\begin{tabular}{|ccccccc|}
\hline
metrics & \makecell[c]{reference\\ model} & \makecell[c]{total\\deletion} & \makecell[c]{correlation\\deletion} & \makecell[c]{fair-SMOTE\\Adaptation} & \makecell[c]{exponentiated\\gradient} & \makecell[c]{fair\\redistribution} \\ \hline
HGR KDE & 28.97\% & 18.25\% & 27.66\% & 28.05\% & 31.74\% & 29.05\% \\
RMSE & 164.21 & 170.61 & 163.03 & 164.88 & 164.30 & 164.72\\
Loss ratio& 99.71\% & 99.30\% & 99.79\% & 99.69\% & 99.70\% & 99.39\% \\ \hline     
\end{tabular}}
\caption{Summary of bias mitigation results : the HGR KDE metric evaluates the dependance between $\hat Y$ and $S$. RMSE ans Loss ratio, defined in (\ref{RMSE_LossRatio}), evaluate the performance of the model after various mitigation strategies.}
\label{tab:recap_mitigation_app}
\end{table}

\section{Conclusion}\label{sec:6}

Fairness serves as a crucial constraint in the realm of pricing. Whether it arises from regulatory requirements or a company's strategic objectives, it is imperative to establish, assess, and alleviate biases to attain equitable models. Consequently, there was an initial examination of fairness from a mathematical standpoint, wherein various metrics were introduced to identify biases both before and after the modeling process, irrespective of the conventional gender-related treatment. Subsequently, a study of mitigation methods was conducted, which encompassed pre-, mid-, and post-modeling interventions. These interventions aimed to address bias comprehensively by reprocessing the data and imposing fairness constraints, as well as to address bias individually by reprocessing distinct premiums. While some methods proved more effective than others, each approach offered valuable insights. 

\dcorr In this application, taking into account the constraint of pricing model performance, unfairness as defined by our metrics was not fully mitigated because it would have led to a significant degradation of our models. For example, in total deletion more variables could be deleted or more redistribution iterations could be made in the fair redistribution method. Apart from this constraint, full fairness seems difficult to achieve, at least because corrections are made on a training dataset and with statistical randomness, changes in the portfolio structure etc. fairness would be partial.\fcorr

In continuation of this research, further datasets and scenarios can be explored to obtain more robust results and establish a reference benchmark. It will also be essential to extend the analysis to encompass other pricing granularities such as portfolio segments or groups of coverages.

This study, nevertheless, lays the groundwork for quantifying and mitigating biases within the insurance domain, providing a framework for ongoing exploration. For instance, there is potential to enhance mitigation techniques during the modeling process and investigate the simultaneous treatment of multiple sensitive variables.

\section*{\normalsize{Acknowledgements}}

\noindent{\bf Data:~} Datasets used in this manuscript are based on confidential data obtained by a consulting company, from a (real) insurance company.\\

\noindent{\bf Conflicts of interest:~} The authors declare no conflicts of interest.\\

\noindent{\bf Declaration of funding:~} 
AC acknowledges Canada's National Sciences and Engineering Research Council (NSERC) for funding (RGPIN-2019-07077) and the SCOR Foundation.\\

\noindent{\bf Acknowledgements:~}
We are grateful to Laurence Barry for stimulating discussions and André Grondin for sharing this thoughts on earlier versions of this work.\\
FV thanks the France 2030 framework programme Centre Henri Lebesgue ANR-11-LABX-0020-01 for its stimulating mathematical research programs.
\\
We are very grateful to the anonymous referees for valuable comments.\\

\newpage

\begin{appendix}

\section{Fair-SMOTE adaptation pseudo code}\label{secA1}

\bigskip
\begin{algorithm}
\caption{Fair-SMOTE adaptation}\label{alg:1}
\label{algo:fair-smote}
\begin{algorithmic}[1]
\For{ each subgroup}
\For{ each simulated individual of the given subgroup}
\State randomly select an individual $p$ in the sub group
\State find via $k$-nearest neighbor the two $p$'s closest neighbors $v_1 \text{ and }v_2$ 
\State sample $u$ from an uniform distribution
\For{each column of our $N$ columns}

 \If {column with binary values} 
        \If {$st>u$}
            \State $\Tilde{x} = \text{\texttt{random\_choice}} (x_{v_1}, x_{v_2}, x_{p})$
        \Else
            \State $\Tilde{x}=x_{p}$
        \EndIf
        
\ElsIf{ column with qualitative values} 
        \State $\Tilde{x} = \text{\texttt{random\_choice}} (x_{v_1}, x_{v_2}, x_{p})$

\ElsIf{column with quantitative values}
        \If {$st>u$}
            \State $\Tilde{x} = x_{p} + ft\times( x_{v_1}- x_{v_2})$
        \Else
            \State $\Tilde{x}=x_{p}$
            \EndIf

\Else
        \State return an alert to trigger the transformation of the column

\EndIf
\EndFor
\EndFor
\EndFor
\end{algorithmic}
\end{algorithm}
\bigskip

With $u$ the realization of an uniform law on $[0,1]$, $\Tilde{x}$ the value of the coordinate of the new individual on the corresponding $j$ column $X_j$, $(x_{j})_{j\in\{v_1,v_2,p\}}$ the value of the coordinate for the individuals $v_1, v_2 \text{ and } p.$ \texttt{Random\_choice}, a random choice between parameters with each parameter having the same chances of being choice. 

\end{appendix}

\bibliographystyle{plainnat}
\bibliography{sn-bibliography}



\end{document}